\documentclass{article}

\usepackage{graphicx,amsmath,amsfonts,amssymb,bm,hyperref,url,epsfig,epsf,color,fullpage,MnSymbol,mathbbol, fmtcount,semtrans,bbm
} 

\usepackage{tabularx}
\usepackage{adjustbox}
\usepackage{booktabs}
\usepackage{placeins}
\usepackage{multicol} 
\usepackage{titlesec}
\usepackage[numbers]{natbib}
\usepackage{pgfplots} 
\pgfplotsset{compat=1.17} 
\usepackage{float}

\usepackage{pgfplots}
\pgfplotsset{compat=1.17}

\usepackage{tikz}
\usepackage{stackengine}

\usepackage{mdframed,lipsum}

\usepgfplotslibrary{groupplots}
\usetikzlibrary{pgfplots.groupplots}

\usepackage{amsmath}
\usepackage{framed}
\usepackage{algorithm}

\usepackage{mdframed,lipsum}

\usepackage[english]{babel}
\usepackage{amsthm}

\newtheorem{theorem}{Theorem}

\newtheorem{proposition}{Proposition}
\newtheorem{definition}{Definition}
\newtheorem{assumption}{Assumption}

\newtheorem{remark}{Remark}

\DeclareMathOperator*{\argmax}{arg\,max}
\DeclareMathOperator*{\argmin}{arg\,min}

\usepackage{graphicx,amsmath,amsfonts,amssymb,bm,hyperref,url,epsfig,epsf,color,fullpage,MnSymbol,mathbbol, fmtcount,algorithmic,algorithm, semtrans
} 

\usepackage{PRIMEarxiv}

\usepackage{subcaption}

\usepackage[utf8]{inputenc} 
\usepackage[T1]{fontenc}    
\usepackage{hyperref}       
\usepackage{url}            
\usepackage{booktabs}       
\usepackage{amsfonts}       
\usepackage{nicefrac}       
\usepackage{microtype}      
\usepackage{lipsum}
\usepackage{fancyhdr}       
\usepackage{graphicx}       
\graphicspath{{media/}}     

\title{Transfer Neyman-Pearson Algorithm for\\ Outlier Detection
}

\author{
  Mohammadreza M. Kalan\\
  Statistics, Columbia University \\
  \texttt{mm6244@columbia.edu} \\
   \And
  Eitan J. Neugut \\
  Statistics, Columbia University\\
  \texttt{eitan.neugut@columbia.edu} \\
  \AND
   Samory Kpotufe\\
  Statistics, Columbia University \\
  \texttt{samory@columbia.edu} \\
}

\begin{document}
\maketitle

\begin{abstract}
We consider the problem of transfer learning in outlier detection where target abnormal data is rare. While transfer learning has been considered extensively in traditional balanced classification, the problem of transfer in outlier detection and more generally in imbalanced classification settings has received less attention. We propose a general meta-algorithm which is shown theoretically to yield strong guarantees w.r.t. to a range of changes in abnormal distribution, and at the same time amenable to practical implementation. We then investigate different instantiations of this general meta-algorithm, e.g., based on multi-layer neural networks, and show empirically that they outperform natural extensions of transfer methods for traditional balanced classification settings (which are the only solutions available at the moment). 
\end{abstract}

\section{Introduction}
Outlier detection problems are characterized by a significant imbalance between two classes of data: one class with an abundance of available samples, referred to as the \textit{common class}, and another with very few or no samples, known as the \textit{outlier} or \textit{rare class}. This imbalance makes it challenging to design accurate decision rules, as the scarcity of data from the rare class hinders the learning process. Examples of applications in this imbalanced setting include detecting rare events in climate science, such as heavy precipitation \citep{folino2023learning,mazzoglio2019improving,frame2022deep}, as well as disease diagnosis \citep{bourzac2014diagnosis,myszczynska2020applications} and malware detection in cybersecurity \citep{alamro2023automated,kumar2019edima}. A proven useful way to address these data limitations is to leverage another related data, referred to as the \textit{source}, which might contain information about the \textit{target} rare class. For instance, in the context of heavy precipitation detection, such related data could come from another location with sufficient recorded samples. This scenario represents a transfer learning problem. However, much of the existing literature has focused on transfer learning in traditional balanced classification tasks \citep{pan2009survey,zhuang}, rather than on outlier detection and imbalanced classification, where there is an asymmetry in the errors across different classes due to their varying relative importance.

In this work, we propose a general meta-algorithm for outlier detection that effectively leverages source task data with sufficient outlier class samples alongside target data. The approach is supported by a theoretical guarantee on the target generalization error, without making any assumptions about the underlying data distribution, while being amenable to practical implementation. Additionally, the proposed meta-algorithm is adaptive, as it exploits source data when it is related to the target and avoids negative transfer when the source is unrelated. Another key feature of the meta-algorithm is its model-free property, enabling it to be applied across a variety of models, such as neural networks, kernel machines, and others. Consequently, this general approach can be integrated with existing methods that use specific models to find a shared representation of feature spaces for the source and target.


To provide a theoretical justification for the performance of the proposed approach, we adopt the transfer Neyman-Pearson framework introduced by \cite{kalantight} to derive generalization error bounds. In the Neyman-Pearson classification problem, the goal is to achieve low classification error on the rare class while ensuring that the error w.r.t. the common class remains below a pre-specified threshold. \cite{kalantight} introduced a transfer Neyman-Pearson framework based on $0$-$1$ loss risk and derived a minimax rate for the problem. 
In this work, we first extend the transfer Neyman-Pearson framework \citep{kalantight} to the case with a surrogate loss function. And then we propose a meta-algorithm as a constrained optimization procedure leveraging source samples along with target samples in outlier detection. Subsequently, we derive a bound on the target generalization error of the solution obtained through the proposed optimization procedure, capturing the extent of information transferable from the source to the target. Furthermore, the bound guarantees that when the source is unrelated to the target, the procedure effectively disregards the source and avoids negative transfer. It is expressed in terms of the number of source and target samples, the Rademacher complexity of the hypothesis class, and a natural extension of the transfer exponent \citep{hanneke2019value}, which quantifies the relative effect of the source on the target. 

We then propose a transfer learning algorithm to implement an instantiation of the proposed theoretically sound optimization procedure. As detailed in Section \ref{sec:transfer_learning_algorithm}, the process begins with constructing the Lagrangian using some tuning parameters. By minimizing the cost over a grid of parameter values, we obtain a function that minimizes the cost for each tuning parameter. Collecting these functions results in a reduced hypothesis class. Subsequently, a function is selected from this reduced class that minimizes the objective function of the proposed optimization procedure while satisfying its constraints. The challenge in transfer learning lies in determining the appropriate bias between the source and target data. Simply optimizing over the target sample may under-utilize valuable information from the source, while solely optimizing over the source samples risks negative transfer if the source distribution diverges significantly from the target. The proposed algorithm addresses this challenge by effectively leveraging the source when it is informative and avoiding negative transfer when the source is unrelated to the target.


We evaluate the proposed algorithm on both real and synthetic datasets. For heavy rainfall prediction, treated as outliers, we use climate data \citep{yu2024climsim,nasa_power_api}, and for default prediction, we use financial data \citep{github_credit}. Our results demonstrate that when the source contains useful information about the target, the algorithm's performance improves compared to using only target data. Conversely, when the source is unrelated, there is no negative transfer effect. In other words, the algorithm adapts to the data and does not require prior knowledge of the relatedness between the source and target. For comparison with other methods, and given the absence of any implementable algorithm for the transfer Neyman-Pearson problem, we propose a practical adaptation of the procedure introduced in \cite{kalantight}. Furthermore, we extend existing baselines—which adjust the scoring function’s threshold to satisfy a pre-specified Type-I error rate \citep{uyar2010handling,abd2013review,saito2015precision}—to the context of transfer learning for outlier detection. Our results demonstrate that the proposed approach consistently achieves superior performance compared to these alternatives.



\section{Related Work}
Unlike outlier detection and imbalanced classification, transfer learning has been widely studied in traditional classification, resulting in the development of numerous approaches and algorithms, as well as the derivation of bounds on the target generalization error. Seminal works in transfer learning, such as \cite{blitzer2007learning, mansour2009domain, ben2010theory, ben2010impossibility}, along with more recent works \citep{zhao2019learning, hanneke2019value, cai2021transfer}, study how knowledge can be transferred from source to target in traditional classification tasks. Particularly relevant to this paper, \cite{hanneke2019value} introduces the concept of the transfer exponent to measure the distance between the source and target domains in traditional classification. We adapt this notion to provide theoretical justification for our proposed transfer learning algorithm.


Outlier detection approaches are generally divided into two categories: semi-supervised and supervised. In the semi-supervised category, where samples are available only from the common or \textit{normal} class, a widely used approach is density level set estimation, which identifies a region of low density to classify outliers \citep{steinwart2005classification, polonik1995measuring, tsybakov1997nonparametric}. \cite{abe2006outlier} reduces the outlier detection problem to a traditional classification task by generating artificial outlier samples. In transfer outlier detection, many studies consider this semi-supervised framework \citep{chalapathy2018anomaly,yang2023anomaly}. For instance, \cite{andrews2016transfer} employs a neural network pre-trained on a supervised task to extract discriminative features of the normal class, followed by a one-class SVM to detect outliers. 

In supervised outlier detection setting, which is also the focus of our work, most algorithms train a scoring function and produce a Receiver Operating Characteristic (ROC) curve by evaluating different thresholds on the scoring function to adjust the type-I error \citep{uyar2010handling,saito2015precision,tong2018neyman}. In contrast, our procedure minimizes the Type-II error for a pre-specified threshold on the Type-I error by effectively leveraging both source and target samples, as detailed in Section \ref{sec:main_results}. Experiments demonstrate that our methods consistently utilize source information when it is relevant and effectively avoid negative transfer when the source is uninformative, without requiring any prior knowledge of the relatedness. This contrasts with other methods, which may perform well in certain scenarios but lack consistent reliability.

More relevant to this work, \cite{kalantight} studied transfer learning in the Neyman-Pearson problem for outlier detection.  \cite{kalantight} demonstrates that outlier detection fundamentally differs from traditional balanced classification, as some seemingly unrelated source and target tasks in classification can still be transferable in the context of outlier detection. \cite{kalantight} characterizes the minimax rate for transfer learning in outlier detection and proposes an adaptive procedure that achieves this rate up to a numerical constant. However, it does not provide an implementable algorithm that effectively leverages source samples alongside target samples. In this work, we adopt the Neyman-Pearson classification framework, first proposing a meta-algorithm with theoretical guarantees as a constrained optimization procedure for efficiently leveraging source samples. We then propose an implementable transfer learning algorithm for outlier detection as an instantiation of the meta-algorithm. Additionally, we compare the performance of our method with an algorithm inspired by \cite{kalantight} and show that ours outperforms it.

\section{Setup}
We begin by setting up the Neyman-Pearson classification framework which formalizes supervised outlier detection and then extend it to the transfer learning setting.
\subsection{Neyman-Pearson Classification}
Let $\mu_0$ and $\mu_1$ represent probability distributions on a measurable space $(\mathcal{X}, \Sigma)$. Additionally, let $\mathcal{H}$ be a hypothesis class consisting of functions $h: \mathcal{X} \rightarrow \mathbb{R}$. For a function $h \in \mathcal{H}$, we predict that data $x\in \mathcal{X}$ is generated by $\mu_1$ if $h(x) \geq 0$, and by $\mu_0$ if $h(x) < 0$. In this paper, we study the setting where there is an abundance of data available from $\mu_0$ and only a few data from $\mu_1$. Therefore, we refer to the classes generated by $\mu_0$ and $\mu_1$ as the \textit{common} class and the \textit{rare} (or \textit{outlier}) class, respectively.
\begin{definition}
Type-I and Type-II errors are defined as $R_{\mu_0}(h)=\mathbb{E}_{\mu_0}\left[\mathbbm{1}\left\{h(X)\geq 0 \right\}\right]$ and $R_{\mu_1}(h)=\mathbb{E}_{\mu_1}\left[\mathbbm{1}\left\{h(X)< 0 \right\}\right]$, respectively, where $\mathbbm{1}$ denotes the indicator function.
\end{definition}
Neyman-Pearson classification aims to minimize the Type-II error while keeping the Type-I error below a pre-specified threshold $\alpha$:
\begin{align}\label{Neyman_Pearson_classification}
   &\underset{h\in \mathcal{H}}{\text{Minimize}}\ R_{\mu_1}(h)\nonumber\\
   &\text{s.t.} \ R_{\mu_0}(h)\leq \alpha
\end{align}
The Neyman-Pearson Lemma \citep{lehmann1986testing}, under some mild assumptions, characterizes the universally optimal solution of \eqref{Neyman_Pearson_classification}—when $\mathcal{H}$ consists of all measurable functions from $\mathcal{X}$ to $\mathbb{R}$—as $h^*_{\alpha}(x) = 2\mathbbm{1}\left\{\frac{p_1}{p_0}(x) \geq \lambda\right\} - 1$, provided there exists a $\lambda$ such that $R_{\mu_0}(h^*_{\alpha}) = \alpha$.

In practical settings, surrogate loss functions are preferred over the indicator loss function because the latter is discontinuous and leads to intractable combinatorial optimization problems. Additionally, surrogate loss functions not only penalize misclassified points but also take into account their distance from the decision boundary, resulting in more robust classifiers \citep{bao2020calibrated}. In this section, we aim to establish the foundation for an implementable transfer learning algorithm for outlier detection. To achieve this, we need to replace the $0$-$1$ loss with a surrogate loss.
\begin{definition}\label{surrogate_loss}A function $\varphi: \mathbb{R}\rightarrow \mathbb{R}^{+}$ is called an L-Lipschitz surrogate loss if it is non-decreasing, $\varphi(0)=1$, satisfies $|\varphi(x)-\varphi(y)|\leq L|x-y|$ for all $x,y\in \mathbb{R}$, and there exists a constant $C>0$ such that for all $h\in \mathcal{H}$ and $x\in \mathcal{X}$ we have $\max\left\{\varphi(h(x)),\varphi(-h(x)) \right\}\leq C$.
\end{definition}
In the following definition, we introduce Type-I and Type-II errors with respect to a surrogate loss.
\begin{definition}
    $\varphi$-Type-I and $\varphi$-Type-II errors are defined as $R_{\varphi,\mu_0}(h)=\mathbb{E}_{\mu_0}\left[\varphi(h(X))\right]$ and $R_{\varphi,\mu_1}(h)=\mathbb{E}_{\mu_1}\left[\varphi(-h(X))\right]$
\end{definition}
Next, we define the Rademacher complexity of a hypothesis class $\mathcal{H}$, which serves as a measure of the class's capacity and controls its complexity.
\begin{definition}[Rademacher Complexity \citep{bartlett2002rademacher}]
    Let $X_1,...,X_n$ be i.i.d. samples drawn from a distribution $\mu$ on $\mathcal{X}$. Define the random variable 
    \begin{align*}
\hat{R}_n(\mathcal{H})=\underset{\bf{\sigma}}{\mathbb{E}}\left[\sup_{h\in \mathcal{H}}|\frac{1}{n}\sum_{i=1}^n \sigma_i h(X_i)| \right],
\end{align*}
where $\sigma_1,...,\sigma_n$ are independent uniform $\{\pm 1\}$-valued random variables. The Rademacher complexity of $\mathcal{H}$ is then defined as $R_n(\mathcal{H})=\mathbb{E} \hat{R}_n(\mathcal{H})$ where the expectation is taken w.r.t. the i.i.d. samples.
\end{definition}

{\begin{assumption}\label{assumption_rademacher}
We assume that $R_n(\mathcal{H})\leq \frac{B_{\mathcal{H}}}{\sqrt{n}}$ for some $B_{\mathcal{H}}$ which characterizes the complexity of $\mathcal{H}$. 
\end{assumption}
\begin{remark}
    If the input features are bounded, most practical hypothesis classes, such as linear regression and neural networks, satisfy Assumption \ref{assumption_rademacher}, provided that the coefficients and weights are bounded \citep{golowich2018size}.
\end{remark}

Neyman-Pearson classification with a surrogate loss $\varphi$ is then formulated as follows:
\begin{align}\label{Neyman_Pearson_classification_surrogate}
   &\underset{h\in \mathcal{H}}{\text{Minimize}}\ R_{\varphi,\mu_1}(h)\nonumber\\
   &\text{s.t.} \ R_{\varphi,\mu_0}(h)\leq \alpha
\end{align}
\subsection{Transfer Learning Setup}
Let $\mu_{1,S}$, $\mu_{1,T}$ denote the distributions of the rare class for the source and target, respectively. We consider the following source and target Neyman-Pearson classification problems with a common distribution $\mu_0$ and surrogate loss $\varphi$:

\begin{minipage}{.35\linewidth}
  \centering
\begin{align}\label{eq_source_neyman_pearson}
    &\underset{h\in \mathcal{H}}{\text{Minimize}} \ R_{\varphi,\mu_{1,S}}(h)\nonumber\\
    &\text{s.t.} \ R_{\varphi,\mu_0}(h)\leq \alpha
  \end{align}
\end{minipage}%
\begin{minipage}{.2\linewidth}
\centering \
\end{minipage}%
\begin{minipage}{.35\linewidth}
  \centering
\begin{align}\label{eq_target_neyman_pearson}
    &\underset{h\in \mathcal{H}}{\text{Minimize}}  \ R_{\varphi,\mu_{1,T}}(h)\nonumber\\
    & \text{s.t.} \ R_{\varphi,\mu_0}(h)\leq \alpha
  \end{align}
\end{minipage}

We denote (not necessarily unique) solutions of \eqref{eq_source_neyman_pearson} and \eqref{eq_target_neyman_pearson} by $h^*_{S,\alpha}$ and $h^*_{T,\alpha}$, respectively.

In practical scenarios, the underlying distributions $\mu_0$, $\mu_{1,S}$, and $\mu_{1,T}$ are unknown and only accessible through samples. We consider a setting where there are $n_0$, $n_S$, and $n_T$ i.i.d. samples available from $\mu_0$, $\mu_{1,S}$, and $\mu_{1,T}$, respectively. The learner then aims to return a hypothesis $\hat{h}\in \mathcal{H}$ that minimizes the \textbf{target excess error}
\begin{align}\label{target-excess-risk}
    \mathcal{E}_{1,T}(\hat{h}):=\max\left\{0,R_{\varphi,\mu_{1,T}}(\hat{h}) -R_{\varphi,\mu_{1,T}}(h^*_{T,\alpha})\right\}
\end{align}
subject to the constraint that $R_{\varphi,\mu_{0}}(\hat{h})\leq \alpha+\epsilon_0$, where a slack $\epsilon_0=\epsilon_0(n_0)$, typically of order $n_0^{-1/2}$, is allowed to deviate from the pre-specified threshold.

Next, we adapt the notion of transfer exponent—used in traditional classification \citep{hanneke2019value} and $0$-$1$ loss Neyman-Pearson classification \citep{kalantight}—to capture the transfer distance between source and target in the setting of Neyman-Pearson classification with a surrogate loss.

\begin{definition}[Transfer Exponent]\label{transfer_exponent}
    Let $S^*_{\alpha}\subset \mathcal{H}$ denote the set of solutions of source problem \eqref{eq_source_neyman_pearson}. We call $\rho(r)>0$ a transfer exponent from source \eqref{eq_source_neyman_pearson} to target \eqref{eq_target_neyman_pearson} under $\mathcal{H}$ if there exist $r, c_{\rho(r)}>0$ such that 
    \begin{align}\label{dist}
        c_{\rho(r)}\cdot \max\bigg\{0,R_{\varphi,\mu_{1,S}}(h)-R_{\varphi,\mu_{1,S}}(h^*_{S,\alpha})\bigg\}
        \geq \max\bigg\{0,R_{\varphi,\mu_{1,T}}(h)-R_{\varphi,\mu_{1,T}}(h^*_{S,\alpha})\bigg\}^{\rho(r)} 
    \end{align}
for all $h \in \mathcal{H} \ \text{with} \ R_{\varphi,\mu_0}(h)\leq \alpha+r$, where $h^*_{S,\alpha}=\underset{h\in S^*_{\alpha}}{\argmax} \  R_{\varphi,\mu_{1,T}}(h)$.
\end{definition}
The transfer exponent reflects how well a function's performance in the source domain translates to its performance in target domain---and thus serves as a measure of how informative the source is about the target. The source is most informative when $\rho$ is small and close to 1, and less informative when $\rho$ is large. 
\section{Main  Theoretical Results}\label{sec:main_results}
In this section, we propose a meta-algorithm as a constrained optimization procedure for Neyman-Pearson classification that aims to find a function minimizing target excess error \eqref{target-excess-risk}, subject to the $\varphi$-Type-I error constraint, by leveraging both source and target data. We then analyze this approach by providing upper bounds on the generalization error of the procedure's solution.

\subsection{Transfer Learning Optimization Procedure}
First, we need to define the empirical counterparts of the surrogate losses as follows:
\begin{align*}
\hat{R}_{\varphi,\mu_0}(h) &= \frac{1}{n_0}\sum_{X_i \sim \mu_0} \varphi(h(X_i)), \quad 
\hat{R}_{\varphi,\mu_{1,T}}(h) = \frac{1}{n_T}\sum_{X_i \sim \mu_{1,T}} \varphi(-h(X_i)), \quad 
\hat{R}_{\varphi,\mu_{1,S}}(h) = \frac{1}{n_S}\sum_{X_i \sim \mu_{1,S}} \varphi(-h(X_i)).
\end{align*}
The following proposition provides a concentration result for empirical errors in hypothesis classes with bounded Rademacher complexities.
\begin{proposition}\label{proposition_rademacher}
Let $\delta>0$ and $\mathcal{H}$ be a hypothesis class satisfying Assumption \ref{assumption_rademacher}. Furthermore, suppose that $\hat{R}_{\varphi, \mu}$ denotes empirical error with respect to $n$ i.i.d. samples drawn from a distribution $\mu$, which could be either $\mu_0$ or $\mu_1$. Then, with probability at least $1-\delta$, we have
$$\underset{h\in \mathcal{H}}{\sup}\ |R_{\varphi,\mu}(h)-\hat{R}_{\varphi,\mu}(h)|\leq \frac{4B_{\mathcal{H}}L+C\sqrt{2\log(2/\delta)}}{\sqrt{n}},$$
where $C$ is defined in Definition \ref{surrogate_loss}.
\end{proposition}
Next, we define $\hat{h}_{T,\alpha+\epsilon_0/2}$ as follows:
\begin{align}\label{h_t}
    \hat{h}_{T,\alpha+\epsilon_0/2}=&\argmin_{h\in \mathcal{H}} \ \hat{R}_{\varphi,\mu_{1,T}}(h)\nonumber\\
    & \text{s.t.} \ \hat{R}_{\varphi,\mu_0}(h)\leq \alpha+\epsilon_0/2
\end{align}
Let $\tilde{C}=8B_{\mathcal{H}}L+2C\sqrt{2\log(2/\delta)}$. We then propose the following optimization procedure to solve problem \eqref{eq_target_neyman_pearson} by utilizing both source and target samples:
\begin{mdframed}
\begin{align}\label{algorithm_main}
    \hat{h}=&\argmin_{h\in \mathcal{H}} \ \hat{R}_{\varphi,\mu_{1,S}}(h)\nonumber\nonumber\\
    &\ \text{s.t.} \ \hat{R}_{\varphi,\mu_{1,T}}(h)\leq \hat{R}_{\varphi,\mu_{1,T}}(\hat{h}_{T,\alpha+\epsilon_0/2})+\frac{2\tilde{C}}{\sqrt{n_T}}\nonumber\\
    & \ \ \ \ \ \ \hat{R}_{\varphi,\mu_0}(h)\leq \alpha+\epsilon_0/2
\end{align}
\end{mdframed}
\subsection{Upper bounds on the Generalization Errors}
The following theorem provides upper bounds on the target excess error in terms of the number of available samples from the source and target, as well as the transfer exponent, which captures the distance between the source and target.
\begin{theorem}\label{theorem_1}
Let $\delta>0$ and $\epsilon_0=\frac{\tilde{C}}{\sqrt{n_0}}$, where $\tilde{C}=8B_{\mathcal{H}}L+2C\sqrt{2\log(2/\delta)}$. Moreover, let $\hat{h}$ be the hypothesis returned by the procedure \eqref{algorithm_main}, and let the transfer exponent be $\rho(r)$ with coefficient $c_{\rho(r)}$ for $r\geq \epsilon_0$. Then, with probability at least $1-3\delta$, the hypothesis $\hat{h}$ satisfies 
\begin{align*}
    &\mathcal{E}_{1,T}(\hat{h})\leq \min\left\{c_{\rho(r)}\cdot( \frac{\tilde{C}}{\sqrt{n_S}})^{1/\rho(r)}+4\cdot\Delta, \frac{4\tilde{C}}{\sqrt{n_T}}\right\}\\
    & R_{\mu_0}(\hat{h})\leq R_{\varphi,\mu_0}(\hat{h})\leq \alpha+\epsilon_0.
\end{align*}
where $\Delta=R_{\varphi,\mu_{1,T}}(h^*_{S,\alpha})-R_{\varphi,\mu_{1,T}}(h^*_{T,\alpha+\epsilon_0})$. Here, $h^*_{T,\alpha+\epsilon_0}$ is the solution to problem \eqref{eq_target_neyman_pearson} with the threshold on the $\varphi$-Type-I error set to $\alpha+\epsilon_0$ instead of $\alpha$. 
\end{theorem}

\begin{remark}\label{remark_2}
$\rho(r)$ captures the relative effectiveness of the source samples in the target domain. The lower the value of $\rho(r)$, the more effective the source samples are. Moreover, source samples are useful only up to a certain accuracy, captured by $\Delta$. To reduce the error further, it becomes necessary to leverage target samples.
\end{remark}

\cite{kalantight} derives a similar bound to Theorem \ref{theorem_1} for the problem of 0-1 loss Neyman-Pearson classification, under the assumption of a finite VC class, except for the term $\Delta$, which is defined there as $R_{\mu_{1,T}}(h^*_{S,\alpha}) - R_{\mu_{1,T}}(h^*_{T,\alpha})$, leading to a sharper bound. Next, we make additional assumptions about the hypothesis class $\mathcal{H}$ and the surrogate loss function $\varphi$ to tighten the upper bound in Theorem \ref{theorem_1}.
{\begin{assumption}\label{assumption_hypothesis}
We assume that $\mathcal{H}$ is a convex class, meaning that for any $\theta \in (0,1)$ and any two hypotheses $h_{1}, h_{2} \in \mathcal{H}$, we have $\theta\cdot h_{1} + (1-\theta)\cdot h_{2} \in \mathcal{H}$.
\end{assumption}

Note that classes such as polynomial regression functions and majority votes over a basis of functions are examples that satisfy Assumption \ref{assumption_hypothesis}. However, a class of neural networks with a fixed architecture is generally not closed under convex combinations. Since the Rademacher complexity of the convex hull of a class is equal to that of the class itself, we can instead consider the convex hull of a neural network class, which is convex.

\begin{theorem}\label{theorem_2}
    Assume the setting of Theorem \ref{theorem_1}. Moreover, suppose that $\mathcal{H}$ satisfies Assumption \ref{assumption_hypothesis} and that $\varphi$ is convex. Furthermore, suppose that the set $\{h\in \mathcal{H}: R_{\varphi,\mu_0}(h)\leq \alpha/2\}$ is nonempty. Then, with probability at least $1 - 3\delta$, the hypothesis $\hat{h}$ returned by the procedure \eqref{algorithm_main} satisfies
\begin{align*}
    &\mathcal{E}_{1,T}(\hat{h}) \leq \min \bigg\{ 
    c_{\rho(r)} \cdot \left( \frac{\tilde{C}}{\sqrt{n_S}} \right)^{1/\rho(r)}+ \frac{C'}{\sqrt{n_0}}  + 4 \cdot \tilde{\Delta},  \frac{4\tilde{C}}{\sqrt{n_T}} \bigg\} \\
    & R_{\mu_0}(\hat{h})\leq R_{\varphi,\mu_0}(\hat{h}) \leq \alpha + \epsilon_0.
\end{align*}
where $\tilde{\Delta}=R_{\varphi,\mu_{1,T}}(h^*_{S,\alpha})-R_{\varphi,\mu_{1,T}}(h^*_{T,\alpha})$ and $C'=\frac{8C\tilde{C}}{\alpha}$.
\end{theorem}
The term \(\frac{C'}{\sqrt{n_0}}\) is negligible because \(n_0\) denotes the number of samples drawn from the common distribution $\mu_0$, from which many samples are available. Therefore, Theorem \ref{theorem_2} provides a sharper bound than Theorem \ref{theorem_1}.

\section{Transfer Learning Algorithm for Outlier Detection}\label{sec:transfer_learning_algorithm}
In this section, we propose a transfer learning Neyman-Pearson (TLNP) algorithm for outlier detection based on the optimization procedure \eqref{algorithm_main}. We evaluate its performance using climate data \citep{yu2024climsim,nasa_power_api}, financial data \citep{github_credit}, and synthetically generated datasets. Additionally, we compare its performance with an algorithm inspired by the procedure proposed in \cite{kalantight}, as well as other approaches. We demonstrate that the proposed algorithm consistently avoids negative transfer when the source is uninformative about the target and effectively leverages an informative source when it is, whereas other approaches may occasionally perform well in specific cases but fail to maintain consistency across different datasets.

The main idea of TLNP algorithm is as follows. First,
we consider the Lagrangian associated with \eqref{algorithm_main} and consider the following cost function, with tuning parameters $\lambda_S,\lambda_0$:
\begin{align}\label{cost_function}
\hat{R}_{\varphi,\mu_{1,T}}(h)+\lambda_S \hat{R}_{\varphi,\mu_{1,S}}(h)+\lambda_0 \hat{R}_{\varphi,\mu_0}(h)
\end{align}
Next, over a grid search of $(\lambda_S, \lambda_0)$, we identify functions within the hypothesis class minimizing the cost function \eqref{cost_function}, thereby obtaining a smaller, filtered hypothesis class. Finally, we solve the $0$-$1$ loss counterpart of \eqref{algorithm_main} within this reduced hypothesis class. Here, we provide a detailed explanation of the TLNP process through the following steps. In the following, $\epsilon_0$ is proportional to $\frac{1}{\sqrt{n_0}}$, where $n_0$ is the number of training data points in the normal class. $\epsilon_0$ is a parameter that the user can select; if the user is conservative regarding Type-I $\alpha$ constraint, it should be chosen to be sufficiently small.

\textbf{Step 1) searching over a grid of $(\lambda_S,\lambda_0)$ pairs:} The TLNP algorithm sets $\lambda_S$ to a fixed point and, for each $\lambda_S$, we start with $\lambda_0$ of 1 and fine-tune $\lambda_0$ until $0$-$1$ loss Type-I error of $h$ belongs to the interval $[\alpha-\epsilon_0/2, \alpha+\epsilon_0/2]$. Since the elements of $\mathcal{H}$ are real-valued functions, we apply the
sign function to set binary classifiers and calculate $\hat{R}_{\mu_0}(\text{sign}(h))$.

We start with $\lambda_S$ fixed to one of $12$ points, $(0, 0.05, 0.1, 0.5, 1, 5, 10, 20, 40, 60, 80, 100)$. For each point $(\lambda_S, \lambda_0)$, we train a new function $h\in \mathcal{H}$. The fine-tuning process works by comparing the Type-I error to the $\alpha \pm \epsilon_0/2$ range. If the Type I error is too high (overshoot), the algorithm increases $\lambda_0$ by multiplying it by $(1+\text{increment factor})$. If the error is too low (undershoot), it decreases $\lambda_0$ by multiplying it by $(1-\text{increment factor})$. The initial increment factor is 0.5. Each time the error flips between overshooting and undershooting, the increment factor is halved, allowing for finer adjustments. Once the Type-I error falls within the range $\alpha \pm \epsilon_0/2$, then we move onto the next $\lambda_S$ in the list.

If fewer than 5 successful tunings have been achieved, the search range is expanded by adding additional $\lambda_S$ values. The process stops when $12$ points successfully converge with an acceptable Type-I error, or when the values of $\lambda_S$ become unreasonably small or large. At the end, we obtain a reduced set of hypothesis class $\hat{\mathcal{H}}$ whose elements satisfy Type-I error constraint.

\textbf{Step 2) Filtering $\hat{\mathcal{H}}$ using the target abnormal data:} We first evaluate $\hat{R}_{\mu_{1,T}}$, which represents the target 0-1 loss Type-II error with respect to the target abnormal training data, for the elements of $\hat{\mathcal{H}}$ obtained in the first step. Let $\hat{h}_T \in \hat{\mathcal{H}}$ be the function that yields the lowest $\hat{R}_{\mu_{1,T}}$, i.e., $\hat{R}_{\mu_{1,T}}(\text{sign}(\hat{h}_T)) = \min_{h \in \hat{\mathcal{H}}} \hat{R}_{\mu_{1,T}}(\text{sign}(h))$. Then, inspired by the constraint in the optimization procedure \eqref{algorithm_main}, we identify the functions that are close to $\hat{h}_T$ in terms of target Type-II error. We use a universal constant $c=0.5$ and define $\hat{\mathcal{H}}_T$ as the set of functions $h\in \hat{\mathcal{H}}$ satisfying the inequality: \begin{align}\label{step_2_thresholding}
\hat{R}_{\mu_{1,T}}(\text{sign}(h))\leq \hat{R}_{\mu_{1,T}}(\text{sign}(\hat{h}_T))+\frac{c}{\sqrt{n_T}}
\end{align}
We demonstrate that this universal constant performs well across all datasets, both real-world and synthetic. Moreover, if users have prior knowledge about the relatedness of the source and target, they can adjust this constant accordingly by either decreasing or increasing it. Furthermore, since the constant serves primarily to upper-bound the variance of errors for a given dataset, we propose a method in Appendix \ref{variance_method_appendix} to estimate this variance and use that instead of the constant.

\textbf{Step 3) Filtering $\hat{\mathcal{H}}_T$ using the source abnormal data:} In this step, we evaluate $\hat{R}_{\mu_{1,S}}$, which represents the source $0$-$1$ loss Type-II error with respect to the source abnormal data, for the elements of $\hat{\mathcal{H}_T}$ obtained in the second step. We then select the function that yields the lowest error as the output of the algorithm. Roughly speaking, in this step, if the source is informative, the algorithm leverages it by minimizing the source error. Conversely, if the source is not informative, all functions in $\hat{\mathcal{H}}_T$ can achieve the rate $\frac{1}{\sqrt{n_T}}$ on the target data, thereby avoiding negative transfer.

We also compare TLNP with other approaches, including a procedure inspired by \cite{kalantight}, as detailed below.

\textbf{1) Transfer learning outlier detection} \citep{kalantight}: While \cite{kalantight} did not propose an implementable algorithm, we draw inspiration from the proposed procedure and implement it as follows. We obtain the solutions to \eqref{eq_source_neyman_pearson} and \eqref{eq_target_neyman_pearson} using a Lagrangian approach and then select the best of two based on evaluation with the target abnormal data. In this approach, the source and target data are handled separately rather than being combined.

\textbf{2) Only target Neyman-Pearson:} This approach is similar to TLNP, except that the source data is not utilized. In other words, we set $\lambda_S = n_S = 0$, thereby eliminating step 3 of the TLNP process. The final output is selected in step 2 by minimizing the target abnormal data. Consequently, this approach serves as a baseline for assessing the benefit of leveraging source data.

\textbf{3) Only source Neyman-Pearson:} This approach is similar to the only target Neyman-Pearson approach, except that the target data is replaced with source data.

\textbf{4) Pooled source and target Neyman-Pearson:} This approach follows the idea of the only target Neyman-Pearson approach, but it pools both source and target data instead of just using target data. In other words, it does not distinguish between the two, treating the source data as if it were the target.

\textbf{5) Only target thresholding traditional classification:} This approach disregards the source data and finds a classifier using a scoring function to classify normal and abnormal data, the same as in traditional balanced classification. It then adjusts the threshold on the scoring function to satisfy the Type-I error constraint.

\textbf{6) Pooled source and target thresholding traditional classification:} This approach follows the idea of only target thresholding approach, except it pools both source and target data, instead of just using target data, without distinguishing between them.

\section{Experiments and Numerical Results}

In this section, we evaluate the proposed algorithm on climate data \citep{yu2024climsim,nasa_power_api}, financial data \citep{github_credit}, and synthetically generated datasets for outlier detection. We analyze various source-target pairs to assess the algorithm's adaptability. When the source is relevant to the target, the algorithm effectively leverages this information. Conversely, if the source is not relevant, it avoids negative transfer, unlike other approaches that often fail to perform consistently and may suffer from negative transfer. Additionally, we implement two instantiations of our algorithm using multi-layer perceptron and quadratic models. Furthermore, in all the experiments, \(n_T\) refers to the target abnormal data, and \(n_S\) refers to the source abnormal data. For the normal class, we use only the data from the target domain.

\subsection{Climate Data (Climsim) Experiments \citep{yu2024climsim}}\label{sec:climate_data_climsim}

\begin{figure}
    \centering
    \includegraphics[width=.38\textwidth, height=.3\textwidth]{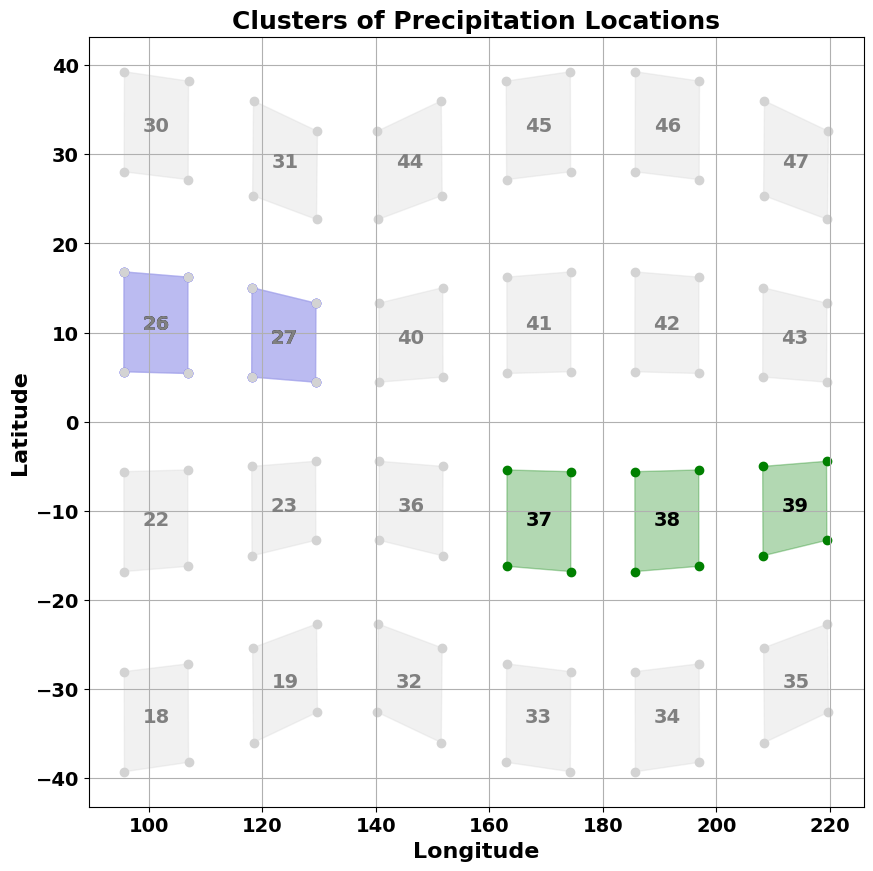}
    \caption{\small Clusters of locations for rain precipitation data \citep{yu2024climsim}, used as source-target pairs. In one scenario, $(26, 27)$ forms a source-target pair, while in another scenario, $38$ is the target, and $37$ and $39$ grouped together constitute the source.} 
    \label{fig:clusters_main} 
\end{figure}

We implement our algorithm, along with the approaches described in Section \ref{sec:transfer_learning_algorithm}, on the ClimSim dataset \citep{yu2024climsim} to detect heavy rain versus non-heavy rain. 

\textbf{Sample Dataset:} In the original dataset, each data point consists of 124 numerical features, such as temperature, specific humidity, and surface sensible heat flux, among others, along with an output of size 128, including variables like rain rate and snow rate. From the output variables, we only extract the rain rate and apply the 95th percentile criterion \citep{saidi2015assessment,schar2016percentile} to categorize the data into binary heavy and non-heavy rain classes. The dataset includes various locations specified by longitude and latitude, which we merge into neighboring clusters. For details on clustering the locations, refer to the Appendix \ref{appendix_D}. Figure \ref{fig:clusters_main} shows a set of location clusters for which we have data on the two rain classes: heavy and non-heavy. We select specific cluster pairs as source and target pairs. In one experiment, we fix the number of target heavy rain samples at \( n_T = 50 \) and increase the number of source heavy rain samples up to 2,500. In another experiment, we fix the number of source heavy rain samples at \( n_S = 2,500 \) and vary \( n_T \) from 25 to 250. In all cases, there are 4,000 training points from the target non-heavy rain class (also referred to as the normal class), along with approximately 2,000 test data points for target heavy rain and 4,000 test data points for target non-heavy rain.

\textbf{Training:} We use a 2-layer fully connected neural network with ReLU activation functions and 62 units in the hidden layer. Additionally, we employ exponential loss as the surrogate loss function and use the Adam optimizer for training. The results are averaged over 10 runs for each experiment.

\textbf{Results:}  In Figures \ref{fig:two_plots_with_shared_legend} and \ref{fig:two_plots_with_shared_legend_38,37,39}, we examine two scenarios: in the first, we select cluster 26 as the target and cluster 27 as the source; in the second, cluster 38 is the target, and clusters 37 and 39 grouped together constitute the source. In these experiments, the Type-I error threshold is set at $\alpha = 0.05$, $\epsilon_0=0.01$, and the Type-II error on the target test data is plotted. Figures \ref{fig:two_plots_with_shared_legend} and \ref{fig:two_plots_with_shared_legend_38,37,39} demonstrate that TLNP effectively combines source and target data to reduce the Type-II error compared to the 'only target' approach, which serves as the baseline. This gain over the baseline, in the case of pairs 26 and 27, is more evident when $n_S$ is sufficiently large. Furthermore, while the "only source NP" and "pooled source and target NP" methods perform relatively well in Figure \ref{fig:two_plots_with_shared_legend_38,37,39} when $n_S$ is sufficiently large, they suffer from negative transfer in Figure \ref{fig:two_plots_with_shared_legend}. A similar pattern is observed with the pooled source and target thresholding method. Although it performs relatively well in Figure \ref{fig:two_plots_with_shared_legend}, its performance is inconsistent, as shown in Figure \ref{fig:two_plots_with_shared_legend_38,37,39}.

Moreover, the results indicate that effectively combining source and target data can outperform even the best of the "only source" and "only target" approaches, including the procedure proposed in \cite{kalantight}. Additionally, Figures \ref{fig:two_plots_with_shared_legend} and \ref{fig:two_plots_with_shared_legend_38,37,39} show that when $n_T$ is fixed at $50$ and $n_S$
increases, the performance of TLNP saturates quickly. This special situation aligns with the scenario described in Remark \ref{remark_2}, where the usefulness of source data quickly saturates, for instance, because the best source predictors differ significantly from the best target predictors (i.e., they have a large error $\Delta$ under the target).

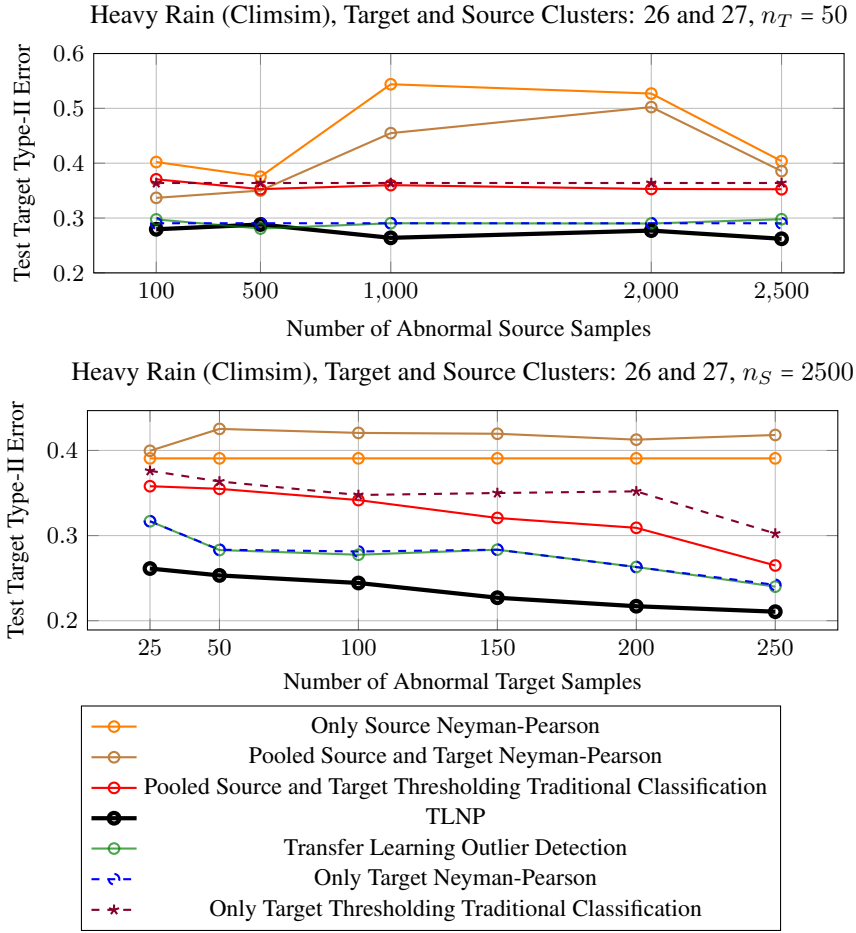
\begin{figure}[h]
    \centering
    \begin{tikzpicture}
        \begin{axis}[
             title={Heavy Rain  (Climsim), Target and Source Clusters: 26 and 27, $n_T=50$},xlabel={\small Number of Abnormal Source Samples},
            ylabel={\small Test Target Type-II Error},
            width=.7\columnwidth, height=4.5cm,
            ymin=0.2, ymax=0.6,
            grid=major,
    xtick={100, 500, 1000, 2000, 2500},ytick={0.2, 0.3, 0.4, 0.5, 0.6},
            legend to name=sharedLegend, 
            legend style={at={(5,-.35)}, anchor=north, legend columns=1, font=\footnotesize},
            label style={font=\small},
            tick label style={font=\footnotesize}
            ]

\addplot[
    mark=o,
    thick,
    color=orange
] coordinates {
    (100, 0.4021)
    (500, 0.3754)
    (1000, 0.54395)
    (2000, 0.527)
    (2500, 0.4038)
};
\addlegendentry{Only Source Neyman-Pearson}

\addplot[
    mark=o,
    thick,
    color=brown
] coordinates {
    (100, 0.3368)
    (500, 0.35011)
    (1000, 0.4548)
    (2000, 0.5023)
    (2500, 0.3856)
};
\addlegendentry{Pooled Source and Target Neyman-Pearson}

\addplot[
    mark=o,
    thick,
    color=red
] coordinates {
    (100, 0.3706)
    (500, 0.3527)
    (1000, 0.3600)
    (2000, 0.3530)
    (2500, 0.3525)
};
\addlegendentry{Pooled Source and Target Thresholding Traditional Classification}

\addplot[
    mark=o,
    ultra thick,
    color=black
] coordinates {
     (100, 0.2795)
    (500, 0.2882)
    (1000, 0.26390)
    (2000, 0.2771)
    (2500, 0.26216)
};
\addlegendentry{Transfer Learning Neyman-Pearson}

\addplot[
    mark=o,
    thick,
    opacity=0.7,
    color=green!50!black,
] coordinates {
    (100, 0.2975)
    (500, 0.2808)
    (1000, 0.2905)
    (2000, 0.290)
    (2500, 0.2978)
};
\addlegendentry{Transfer Learning Outlier Detection}

\addplot[
    mark=o,
    thick,
    dashed,
    color=blue
] coordinates {
    (100, 0.2905)
    (500, 0.2905)
    (1000, 0.2905)
    (2000, 0.2905)
    (2500, 0.2905)
};
\addlegendentry{Only Target Neyman-Pearson}
\addplot[
    mark=star,
    thick,
    dashed,
    color=purple!70!black
] coordinates {
    (100, 0.36390)
    (500, 0.36390)
    (1000, 0.36390)
    (2000, 0.36390)
    (2500, 0.36390)
};
\addlegendentry{Only Target Thresholding Traditional Classification}

        \end{axis}
    \end{tikzpicture}

    \begin{tikzpicture}
        \begin{axis}[
        title={Heavy Rain (Climsim), Target and Source Clusters: 26 and 27, $n_S=2500$},xlabel={\small Number of Abnormal Target Samples},
            ylabel={\small Test Target Type-II Error},
            width=.7\columnwidth, height=4.5cm, 
            grid=major,
            xtick={25, 50, 100, 150, 200,250},
            legend to name=sharedLegend, 
            legend style={at={(5,-.35)}, anchor=north, legend columns=1, font=\footnotesize},
            label style={font=\small},
            tick label style={font=\footnotesize}
            ]

\addplot[
    mark=o,
    thick,
    color=orange
] coordinates {
    (25, 0.3907)
    (50, 0.3907)
    (100, 0.3907)
    (150, 0.3907)
    (200, 0.3907)
    (250, 0.3907)
};
\addlegendentry{Only Source Neyman-Pearson}

\addplot[
    mark=o,
    thick,
    color=brown
] coordinates {
    (25, 0.39957)
    (50, 0.425316)
    (100, 0.4206)
    (150, 0.4196)
    (200,  0.41268)
    (250, 0.418158)
};
\addlegendentry{Pooled Source and Target Neyman-Pearson}

\addplot[
    mark=o,
    thick,
    color=red
] coordinates {
    (25, 0.358)
    (50, 0.3549)
    (100, 0.3417)
    (150, 0.320684)
    (200, 0.30910)
    (250, 0.26489)
};
\addlegendentry{Pooled Source and Target Thresholding Traditional Classification}

\addplot[
    mark=o,
    ultra thick,
    color=black
] coordinates {
    (25, 0.2613)
    (50, 0.25321)
    (100, 0.24431)
    (150, 0.227)
    (200, 0.217)
    (250, 0.2106)
};
\addlegendentry{TLNP}

\addplot[
    mark=o,
    thick,
    opacity=0.7,
    color=green!50!black
] coordinates {
    (25, 0.3168)
    (50, 0.283)
    (100, 0.2775)
    (150, 0.2834)
    (200, 0.2631)
    (250, 0.240)
};
\addlegendentry{Transfer Learning Outlier Detection}

\addplot[
    mark=o,
    thick,
    dashed,
    color=blue
] coordinates {
    (25, 0.316842)
    (50, 0.2834)
    (100, 0.2813)
    (150, 0.283421)
    (200, 0.263)
    (250, 0.2421)
};
\addlegendentry{Only Target Neyman-Pearson}

\addplot[
    mark=star,
    thick,
    dashed,
    color=purple!70!black
] coordinates {
    (25, 0.3761)
    (50, 0.36357)
    (100, 0.347684)
    (150, 0.35)
    (200, 0.351947)
    (250, 0.3024)
};
\addlegendentry{Only Target Thresholding Traditional Classification}

        \end{axis}
    \end{tikzpicture}

    \centering
    \pgfplotslegendfromname{sharedLegend}
    
    \caption{\small The performance of our algorithm (TLNP), along with other approaches on the Climate data \citep{yu2024climsim}, is evaluated for a Type-I error rate of $\alpha=0.05$. In this experiment, one scenario fixes the number of target heavy rain samples at $n_T=50$ while increasing the number of source heavy rain samples $n_S$. In the other scenario, $n_S$ is fixed at $2500$, and $n_T$ is varied. In both cases, the target non-heavy rain class contains 4000 training samples.}
    \label{fig:two_plots_with_shared_legend} 
\end{figure}

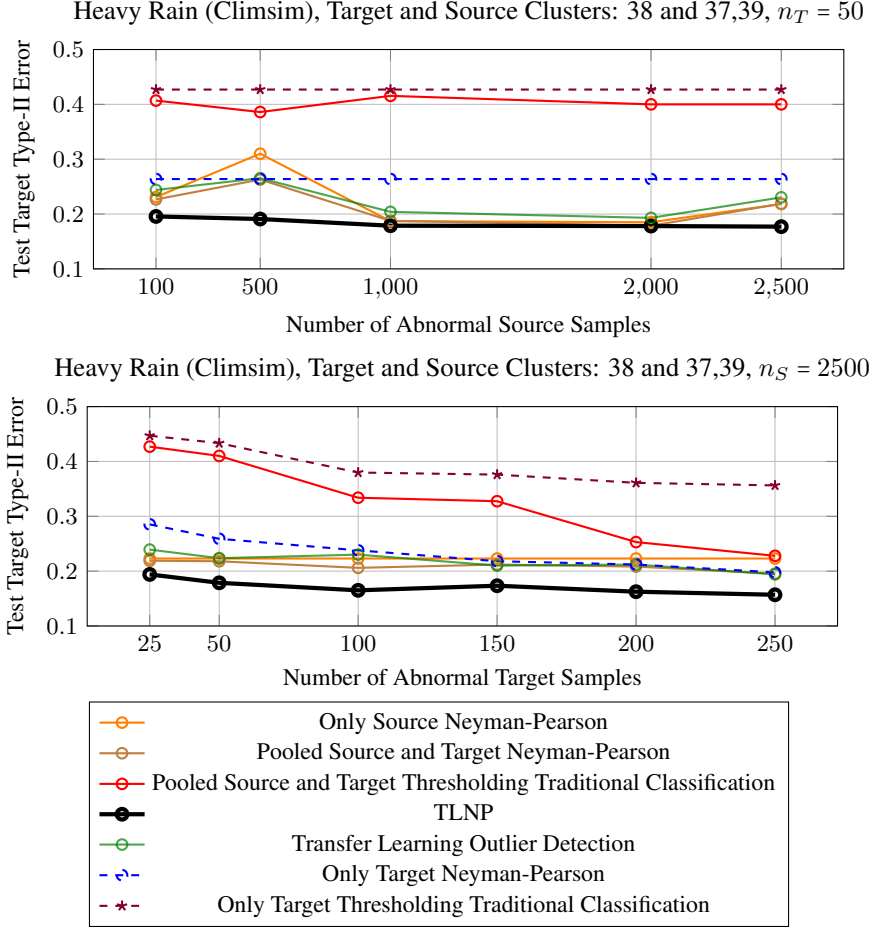
\begin{figure}[h]
    \centering
    \begin{tikzpicture}
        \begin{axis}[
             title={Heavy Rain  (Climsim), Target and Source Clusters: 38 and 37,39, $n_T=50$ },xlabel={Number of Abnormal Source Samples},
            ylabel={Test Target Type-II Error},
            width=.7\columnwidth, height=4.5cm,
            ymin=0.1, ymax=0.5,
            grid=major,
    xtick={100, 500, 1000, 2000, 2500},ytick={0.1,0.2, 0.3, 0.4, 0.5, 0.6},
            legend to name=sharedLegend, 
            legend style={at={(5,-.35)}, anchor=north, legend columns=1, font=\footnotesize},
            label style={font=\small},
            tick label style={font=\footnotesize}
            ]

\addplot[
    mark=o,
    thick,
    color=orange
] coordinates {
    (100, 0.231)
    (500, 0.310)
    (1000, 0.187)
    (2000, 0.185)
    (2500, 0.21827)
};
\addlegendentry{Only Source Neyman-Pearson}

\addplot[
    mark=o,
    thick,
    color=brown
] coordinates {
    (100, 0.22662)
    (500, 0.2629)
    (1000, 0.1876)
    (2000, 0.17864)
    (2500, 0.2188)
};
\addlegendentry{Pooled Source and Target Neyman-Pearson}

\addplot[
    mark=o,
    thick,
    color=red
] coordinates {
    (100, 0.4069)
    (500, 0.386)
    (1000, 0.4155)
    (2000, 0.400)
    (2500, 0.400)
};
\addlegendentry{Pooled Source and Target Thresholding Traditional Classification}

\addplot[
    mark=o,
    ultra thick,
    color=black
] coordinates {
     (100, 0.19559)
    (500, 0.1909)
    (1000, 0.1787)
    (2000, 0.1781)
    (2500, 0.177)
};
\addlegendentry{Transfer Learning Neyman-Pearson}

\addplot[
    mark=o,
    thick,
    opacity=0.7,
    color=green!50!black,
] coordinates {
    (100, 0.2439)
    (500, 0.265356)
    (1000, 0.204)
    (2000, 0.193)
    (2500, 0.2304)
};
\addlegendentry{Transfer Learning Outlier Detection}

\addplot[
    mark=o,
    thick,
    dashed,
    color=blue
] coordinates {
    (100, 0.2638)
    (500, 0.2638)
    (1000, 0.2638)
    (2000, 0.2638)
    (2500, 0.2638)
};
\addlegendentry{Only Target Neyman-Pearson}
\addplot[
    mark=star,
    thick,
    dashed,
    color=purple!70!black
] coordinates {
    (100, 0.427)
    (500, 0.427)
    (1000, 0.427)
    (2000, 0.427)
    (2500, 0.427)
};
\addlegendentry{Only Target Thresholding Traditional Classification}

        \end{axis}
    \end{tikzpicture}

    \begin{tikzpicture}
        \begin{axis}[
        title={Heavy Rain  (Climsim), Target and Source Clusters: 38 and 37,39, $n_S=2500$},xlabel={Number of Abnormal Target Samples},
            ylabel={Test Target Type-II Error},
            width=.7\columnwidth, height=4.5cm, 
            grid=major,
            ymin=0.1, ymax=0.5,
            xtick={25, 50, 100, 150, 200,250},
            legend to name=sharedLegend, 
            legend style={at={(5,-.35)}, anchor=north, legend columns=1, font=\footnotesize},
            label style={font=\small},
            tick label style={font=\footnotesize}
            ]

\addplot[
    mark=o,
    thick,
    color=orange
] coordinates {
    (25, 0.223)
    (50, 0.223)
    (100, 0.223)
    (150, 0.223)
    (200, 0.223)
    (250, 0.223)
};
\addlegendentry{Only Source Neyman-Pearson}

\addplot[
    mark=o,
    thick,
    color=brown
] coordinates {
    (25, 0.219)
    (50, 0.218)
    (100, 0.206)
    (150, 0.212)
    (200,  0.2083)
    (250, 0.19539)
};
\addlegendentry{Pooled Source and Target Neyman-Pearson}

\addplot[
    mark=o,
    thick,
    color=red
] coordinates {
    (25, 0.427)
    (50, 0.4101)
    (100, 0.333879)
    (150, 0.32751)
    (200, 0.253)
    (250, 0.2279)
};
\addlegendentry{Pooled Source and Target Thresholding Traditional Classification}

\addplot[
    mark=o,
    ultra thick,
    color=black
] coordinates {
    (25, 0.194)
    (50, 0.1787)
    (100, 0.165)
    (150, 0.1734)
    (200, 0.1625)
    (250, 0.156909)
};
\addlegendentry{TLNP}

\addplot[
    mark=o,
    thick,
    opacity=0.7,
    color=green!50!black
] coordinates {
    (25, 0.23927)
    (50, 0.2236)
    (100, 0.2301)
    (150, 0.2101)
    (200, 0.212)
    (250, 0.1941)
};
\addlegendentry{Transfer Learning Outlier Detection}

\addplot[
    mark=o,
    thick,
    dashed,
    color=blue
] coordinates {
    (25, 0.28527)
    (50, 0.2590)
    (100, 0.2379)
    (150, 0.2181)
    (200, 0.212)
    (250, 0.19781)
};
\addlegendentry{Only Target Neyman-Pearson}

\addplot[
    mark=star,
    thick,
    dashed,
    color=purple!70!black
] coordinates {
    (25, 0.4470)
    (50, 0.43333)
    (100, 0.3798)
    (150, 0.376)
    (200, 0.361)
    (250, 0.3563)
};
\addlegendentry{Only Target Thresholding Traditional Classification}

        \end{axis}
    \end{tikzpicture}

    \centering
    \pgfplotslegendfromname{sharedLegend}
    
    \caption{\small The performance of our algorithm (TLNP), along with other approaches on the Climate data \citep{yu2024climsim}, is evaluated for a Type-I error rate of $\alpha=0.05$. In this experiment, one scenario fixes the number of target heavy rain samples at $n_T=50$ while increasing the number of source heavy rain samples $n_S$. In the other scenario, $n_S$ is fixed at $2500$, and $n_T$ is varied. In both cases, the target non-heavy rain class contains 4000 training samples.}
    \label{fig:two_plots_with_shared_legend_38,37,39} 
\end{figure}


\subsection{NASA Climate Data Experiments \citep{nasa_power_api}}\label{sec:NASA Climate Data}

We use the NASA dataset \citep{nasa_power_api} for heavy rain detection, with target and source locations in the U.S. and Africa.

\textbf{Sample Dataset:} Each data point consists of six numerical features, and the 90th percentile criterion \cite{kim2023exposure} is applied to classify the data into binary categories: heavy rain and non-heavy rain. In one experiment, the number of target heavy rain samples is fixed at \( n_T = 50 \), while the number of source heavy rain samples is increased up to 2,500. In another experiment, the number of source heavy rain samples is fixed at \( n_S = 2,500 \), and \( n_T \) is varied from 25 to 250. Additionally, in all scenarios, there are approximately 4,000 points from the target non-heavy rain class, while the test set includes 2,000 points for target heavy rain and 5,000 points for target non-heavy rain.

\textbf{Training:} We utilize a two-layer fully connected neural network with ReLU activation functions and 12 units in the hidden layer. The exponential loss function is employed as a surrogate loss, and training is conducted using the Adam optimizer. Results are averaged over 10 runs for each experiment.

\textbf{Results:} Since these locations are geographically distant (e.g., the U.S. and Africa), the source data is not expected to be related to the target. Figure \ref{fig:two_plots_with_shared_legend_NASA} demonstrates that our proposed algorithm effectively avoids negative transfer and achieves performance comparable to the baseline that uses only the target data. In contrast, irrelevant source data negatively impacts the performance of other methods that incorporate source data in a naive manner. In this experiment, the Type-I error threshold is set to $\alpha=0.05$, with $\epsilon_0=0.01$.

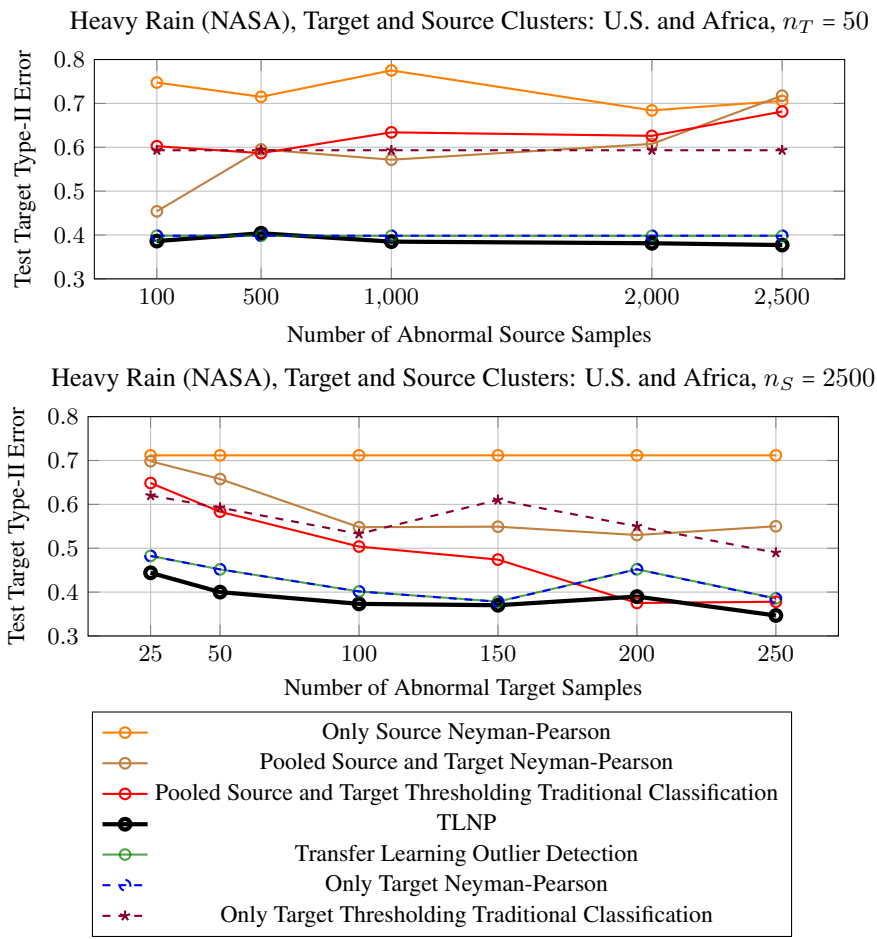
\begin{figure}[h]
    \centering
    \begin{tikzpicture}
        \begin{axis}[
             title={Heavy Rain (NASA), Target and Source Clusters: U.S. and Africa, $n_T=50$},xlabel={Number of Abnormal Source Samples},
            ylabel={Test Target Type-II Error},
            width=.7\columnwidth, height=4.5cm,
            ymin=0.3, ymax=0.8,
            grid=major,
    xtick={100, 500, 1000, 2000, 2500},ytick={0.3,0.4, 0.5, 0.6, 0.7, 0.8},
            legend to name=sharedLegend, 
            legend style={at={(5,-.35)}, anchor=north, legend columns=1, font=\footnotesize},
            label style={font=\small},
            tick label style={font=\footnotesize}
            ]

\addplot[
    mark=o,
    thick,
    color=orange
] coordinates {
    (100, 0.7475)
    (500, 0.715)
    (1000, 0.7752)
    (2000, 0.684)
    (2500, 0.7056)
};
\addlegendentry{Only Source Neyman-Pearson}

\addplot[
    mark=o,
    thick,
    color=brown
] coordinates {
    (100, 0.454)
    (500, 0.5951)
    (1000, 0.5716)
    (2000, 0.6076)
    (2500, 0.7177)
};
\addlegendentry{Pooled Source and Target Neyman-Pearson}

\addplot[
    mark=o,
    thick,
    color=red
] coordinates {
    (100, 0.6024)
    (500, 0.5866)
    (1000, 0.634)
    (2000, 0.626)
    (2500, 0.6815)
};
\addlegendentry{Pooled Source and Target Thresholding Traditional Classification}

\addplot[
    mark=o,
    ultra thick,
    color=black
] coordinates {
     (100, 0.3862)
    (500, 0.40395)
    (1000, 0.3849)
    (2000, 0.3812)
    (2500, 0.3771)
};
\addlegendentry{Transfer Learning Neyman-Pearson}

\addplot[
    mark=o,
    thick,
    opacity=0.7,
    color=green!50!black,
] coordinates {
    (100, 0.39835)
    (500, 0.3983)
    (1000, 0.398)
    (2000, 0.398)
    (2500, 0.398)
};
\addlegendentry{Transfer Learning Outlier Detection}

\addplot[
    mark=o,
    thick,
    dashed,
    color=blue
] coordinates {
    (100, 0.39835)
    (500, 0.39835)
    (1000, 0.39835)
    (2000, 0.39835)
    (2500, 0.39835)
};
\addlegendentry{Only Target Neyman-Pearson}
\addplot[
    mark=star,
    thick,
    dashed,
    color=purple!70!black
] coordinates {
    (100, 0.5932)
    (500, 0.5932)
    (1000, 0.5932)
    (2000, 0.5932)
    (2500, 0.5932)
};
\addlegendentry{Only Target Thresholding Traditional Classification}

        \end{axis}
    \end{tikzpicture}

    \begin{tikzpicture}
        \begin{axis}[
        title={Heavy Rain (NASA), Target and Source Clusters: U.S. and Africa, $n_S=2500$},xlabel={Number of Abnormal Target Samples},
            ylabel={Test Target Type-II Error},
            width=.7\columnwidth, height=4.5cm, 
            grid=major,
            ymin=0.3, ymax=0.8,
            xtick={25, 50, 100, 150, 200,250}, ytick={0.3,0.4,0.5,0.6,0.7,0.8},
            legend to name=sharedLegend, 
            legend style={at={(5,-.35)}, anchor=north, legend columns=1, font=\footnotesize},
            label style={font=\small},
            tick label style={font=\footnotesize}
            ]

\addplot[
    mark=o,
    thick,
    color=orange
] coordinates {
    (25, 0.7117)
    (50, 0.7117)
    (100, 0.7117)
    (150, 0.7117)
    (200, 0.7117)
    (250, 0.7117)
};
\addlegendentry{Only Source Neyman-Pearson}

\addplot[
    mark=o,
    thick,
    color=brown
] coordinates {
    (25, 0.6983)
    (50, 0.6577)
    (100, 0.548)
    (150, 0.54912)
    (200,  0.5301)
    (250, 0.5501)
};
\addlegendentry{Pooled Source and Target Neyman-Pearson}

\addplot[
    mark=o,
    thick,
    color=red
] coordinates {
    (25, 0.6486)
    (50, 0.58328)
    (100, 0.503795)
    (150, 0.4741)
    (200, 0.37523)
    (250, 0.378)
};
\addlegendentry{Pooled Source and Target Thresholding Traditional Classification}

\addplot[
    mark=o,
    ultra thick,
    color=black
] coordinates {
    (25, 0.444)
    (50, 0.400)
    (100, 0.3732)
    (150, 0.3701)
    (200, 0.3898)
    (250, 0.3467)
};
\addlegendentry{TLNP}

\addplot[
    mark=o,
    thick,
    opacity=0.7,
    color=green!50!black
] coordinates {
    (25, 0.48246)
    (50, 0.4518)
    (100, 0.4015)
    (150, 0.37846)
    (200, 0.452)
    (250, 0.3859)
};
\addlegendentry{Transfer Learning Outlier Detection}

\addplot[
    mark=o,
    thick,
    dashed,
    color=blue
] coordinates {
    (25, 0.4824)
    (50, 0.4518)
    (100, 0.4015)
    (150, 0.378)
    (200, 0.452)
    (250, 0.385)
};
\addlegendentry{Only Target Neyman-Pearson}

\addplot[
    mark=star,
    thick,
    dashed,
    color=purple!70!black
] coordinates {
    (25, 0.620)
    (50, 0.5925)
    (100, 0.5328)
    (150, 0.610)
    (200, 0.550)
    (250, 0.4898)
};
\addlegendentry{Only Target Thresholding Traditional Classification}

        \end{axis}
    \end{tikzpicture}

    \centering
    \pgfplotslegendfromname{sharedLegend}
    
    \caption{\small The performance of our algorithm (TLNP), along with other approaches on the Climate data \citep{yu2024climsim}, is evaluated for a Type-I error rate of $\alpha=0.05$. In this experiment, one scenario fixes the number of target heavy rain samples at $n_T=50$ while increasing the number of source heavy rain samples $n_S$. In the other scenario, $n_S$ is fixed at $2500$, and $n_T$ is varied. In both cases, the target non-heavy rain class contains 4000 training samples.}
    \label{fig:two_plots_with_shared_legend_NASA} 
\end{figure}

\subsection{Financial Data Experiments \citep{github_credit}}
In this dataset, the goal is to predict whether a person will become financially delinquent within two years, meaning they fail to repay an installment that is 90 days or more past due.

\textbf{Sample Dataset:} We group the data based on age, with individuals 36 years old and younger as the target group, and those 37 and older (where there is substantially more data) as the source group. The dataset contains nine input features, including personal credit balance, monthly income, debt-to-income ratio, and the number of late payments. In the experiment, we fix the number of source abnormal samples at \( n_S = 2500 \) and vary the number of target abnormal samples \( n_T \) from 25 to 250. Additionally, there are 4,000 points from the target normal class for training, while the test set contains around 2,000 points from the target abnormal class and 5,000 points from the target normal class.

\textbf{Training:} We use a two-layer fully connected neural network with ReLU activation functions and $9$ units in the hidden layer. The exponential loss function is used as a surrogate loss, and training is performed with the Adam optimizer. Results are averaged over 10 runs for each experiment.

\textbf{Results:} Figure \ref{fig:credit} illustrates that our algorithm (TLNP) effectively leverages source information, achieving a notable reduction in Type-II error compared to the baseline 'only target' approach. Additionally, TLNP outperforms other methods by efficiently integrating both source and target data. Furthermore, the results indicate that while naively using source data does not yield good performance on the target, effectively combining source and target data can lead to significant improvements. Here, the threshold on Type-I error is set at $\alpha=0.1$, with $\epsilon_0=0.01$. 

\begin{figure}[h]
    \centering
    \begin{tikzpicture}
        \begin{axis}[
            title={Financial Data, $n_S=2500$},xlabel={Number of Abnormal Target Samples},
            ylabel={Test Target Type-II Error},
            legend pos=north west,
            grid=major,
            xtick={25, 50, 100, 150, 200,250},
            label style={font=\small},
            width=.7\columnwidth, height=4.5cm,
            legend style={at={(0.5,-.45)}, anchor=north, legend columns=1, font=\footnotesize},
            ]

\addplot[
    mark=o,
    thick,
    color=orange
] coordinates {
    (25, 0.829)
    (50, 0.829)
    (100, 0.829)
    (150, 0.829)
    (200, 0.829)
    (250, 0.829)
};
\addlegendentry{Only Source Neyman-Pearson}

\addplot[
    mark=o,
    thick,
    color=brown
] coordinates {
    (25, 0.8073)
    (50, 0.838)
    (100, 0.86884)
    (150, 0.8774)
    (200, 0.864)
    (250, 0.829)
};
\addlegendentry{Pooled Source and Target Neyman-Pearson}

\addplot[
    mark=o,
    thick,
    color=red
] coordinates {
    (25, 0.841)
    (50, 0.842)
    (100, 0.8345)
    (150, 0.8330)
    (200, 0.8274)
    (250, 0.8267)
};
\addlegendentry{Pooled Source and Target Thresholding Traditional Classification}

\addplot[
    mark=o,
    ultra thick,
    color=black
] coordinates {
    (25, 0.5906)
    (50, 0.5728)
    (100, 0.576)
    (150, 0.568)
    (200, 0.546)
    (250, 0.525)
};
\addlegendentry{TLNP}

\addplot[
    mark=o,
    thick,
    opacity=0.7,
    color=green!50!black
] coordinates {
    (25, 0.8286)
    (50, 0.8207)
    (100, 0.805)
    (150, 0.788)
    (200, 0.7121)
    (250, 0.6158)
};
\addlegendentry{Transfer Learning Outlier Detection}

\addplot[
    mark=o,
    thick,
    dashed,
    color=blue
] coordinates {
    (25, 0.843)
    (50, 0.8336)
    (100, 0.8234)
    (150, 0.801)
    (200, 0.7170)
    (250, 0.615)
};
\addlegendentry{Only Target Neyman-Pearson}

\addplot[
    mark=star,
    thick,
    dashed,
    color=purple!70!black
] coordinates {
    (25, 0.8432)
    (50, 0.842)
    (100, 0.8361)
    (150, 0.83342)
    (200, 0.8317)
    (250, 0.8308)
};
\addlegendentry{Only Target Thresholding Traditional Classification}

        \end{axis}
    \end{tikzpicture}
    \caption{\small The performance of our algorithm (TLNP), along with other approaches, on financial data \citep{github_credit} for predicting whether a person will become financially delinquent. The threshold on Type-I error is set at $\alpha=0.1$. In this experiment, the number of source samples is fixed at $n_S=2500$, and $n_T$ is varied from 25 to 250. Moreover, the target normal class contains 4000 training samples.}
    \label{fig:credit} 
\end{figure}
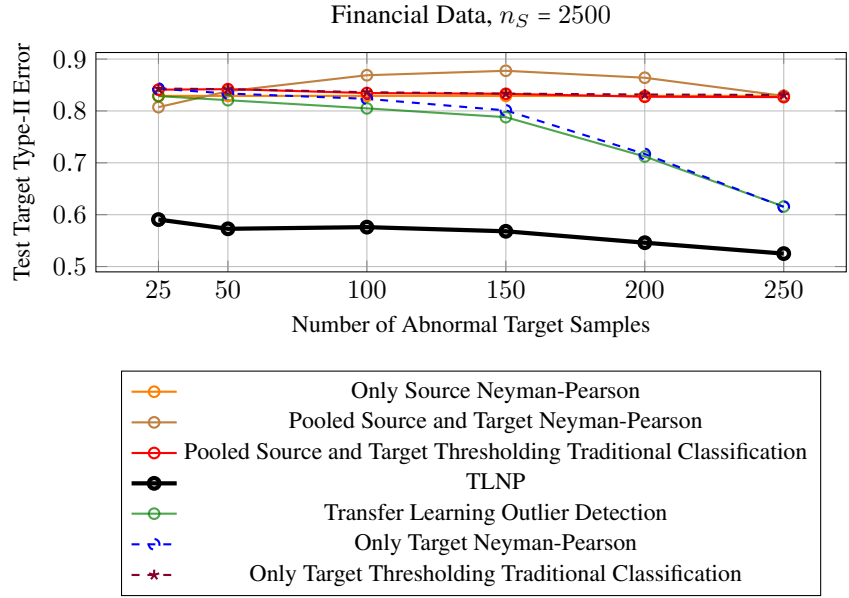

\subsection{Synthetic Data Experiments}
In this section, we evaluate the performance of our algorithm alongside other approaches on Gaussian data. Since it is unlikely to have highly similar source and target distributions in real datasets, we investigate this scenario using synthetic data. Furthermore, in this experiment, we use another instantiation of our algorithm with quadratic models.

\textbf{Sample Dataset:} We generate three datasets corresponding to the normal class, target abnormal class, and source abnormal class, each sampled from standard Gaussian distributions with means \(0\), \(0.5\), and \(0.5\), respectively, and a covariance matrix \(I_{15}\), where the number of features is 15. In this case, the source and target distributions are exactly the same. For the target, we generate 4,000 training data points for the normal class and $n_T=50$ for the abnormal class. The number of source abnormal points, \(n_S\), is varied between 100 and 2,500.

\textbf{Training:} We use a quadratic model, $x^T\mathbf{A}x + \mathbf{b}^Tx + c$, where $\mathbf{A}, \mathbf{b},$ and $c$ are the parameters to be learned. Additionally, we employ exponential loss as the surrogate loss function and use the Adam optimizer for training. The results are averaged over 10 runs for each experiment, with new data generated for each run.

\textbf{Results:} The Type-I error threshold is set to \(\alpha = 0.05\), with \(\epsilon_0 = 0.01\). In this case, since the source and target distributions are exactly the same, methods that naively use the source data are expected to perform very well. Figure \ref{fig:synthetic_Gaussian_very_close_source} shows that TLNP achieves performance very close to "only source NP" and "pooled source and target NP," both of which achieve the lowest Type-II errors. However, as stated earlier, the advantage of TLNP is its adaptability and consistent performance, regardless of whether the source is related to the target, without requiring prior knowledge of this relationship. In contrast, methods that naively use the source data perform well only when the source distribution is very similar to the target.

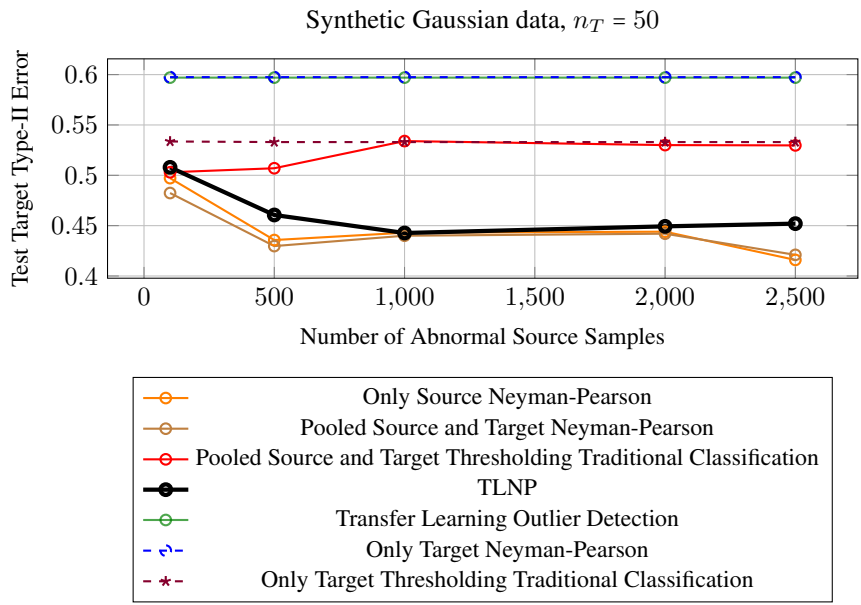
\begin{figure}[h]
    \centering
    \begin{tikzpicture}
        \begin{axis}[
            title={Synthetic Gaussian data, $n_T=50$},xlabel={Number of Abnormal Source Samples},
            ylabel={Test Target Type-II Error},
            legend pos=north west,
            grid=major,label style={font=\small},
            width=.7\columnwidth, height=4.5cm,
            legend style={at={(0.5,-.45)}, anchor=north, legend columns=1, font=\footnotesize},
            ]
            
\addplot[
    mark=o,
    thick,
    color=orange
] coordinates {
    (100, 0.49718)
    (500, 0.4357)
    (1000, 0.443)
    (2000, 0.444)
    (2500, 0.41596)
};
\addlegendentry{Only Source Neyman-Pearson}

\addplot[
    mark=o,
    thick,
    color=brown
] coordinates {
    (100, 0.4824)
    (500, 0.4297)
    (1000, 0.440)
    (2000, 0.442)
    (2500, 0.421)
};
\addlegendentry{Pooled Source and Target Neyman-Pearson}

\addplot[
    mark=o,
    thick,
    color=red
] coordinates {
    (100, 0.503)
    (500, 0.507)
    (1000, 0.534)
    (2000, 0.530)
    (2500, 0.5297)
};
\addlegendentry{Pooled Source and Target Thresholding Traditional Classification}

\addplot[
    mark=o,
    ultra thick,
    color=black
] coordinates {
    (100, 0.5079)
    (500, 0.4606)
    (1000, 0.4427)
    (2000, 0.4493 )
    (2500, 0.452)
};
\addlegendentry{TLNP}

\addplot[
    mark=o,
    thick,
    opacity=0.7,
    color=green!50!black
] coordinates {
    (100, 0.597)
    (500, 0.597)
    (1000, 0.597)
    (2000, 0.597)
    (2500, 0.597)
};
\addlegendentry{Transfer Learning Outlier Detection}

\addplot[
    mark=o,
    thick,
    dashed,
    color=blue
] coordinates {
    (100, 0.5975)
    (500, 0.5975)
    (1000, 0.5975)
    (2000, 0.5975)
    (2500, 0.5975)
};
\addlegendentry{Only Target Neyman-Pearson}

\addplot[
    mark=star,
    thick,
    dashed,
    color=purple!70!black
] coordinates {
    (100, 0.5336)
    (500, 0.533)
    (1000, 0.533)
    (2000, 0.533)
    (2500, 0.533)
};
\addlegendentry{Only Target Thresholding Traditional Classification}

        \end{axis}
    \end{tikzpicture}
    \caption{\small The performance of our algorithm (TLNP), along with other approaches on Gaussian data. The threshold on Type-I error is set at $\alpha=0.05$. The data consists of three sets: the normal class, the target abnormal class, and the source abnormal class. These are generated according to standard Gaussian distributions with means of \(0\), \(0.5\), and \(0.5\), respectively, and a covariance matrix of \(I_{15}\), where the number of features is 15. Furthermore, the  normal class
contains 4000 training samples.
}
    \label{fig:synthetic_Gaussian_very_close_source} 
\end{figure}

\subsection{Overall Performance Summary}

Table \ref{tab:blank_table} summarizes the Type-II errors of various approaches across all datasets for the case where \( n_T = 50 \) and \( n_S = 2500 \). It highlights that the Type-II error is consistently close to the minimum—being the minimum in four datasets and near the minimum in one dataset—indicating adaptability across datasets. In contrast, the performance of each baseline method varies significantly across datasets.

\begin{table}[h]
\centering
\adjustbox{max width=\textwidth}{
\begin{tabularx}{\textwidth}{|X|X|X|X|X|X|}
\hline
\text{Approach} & \text{Climsim; 26} & \text{Climsim; 38} & \text{NASA} & \text{Financial Data} & \text{Gaussian} \\
\hline
TLNP& $\bf{0.26 \pm 0.03}$ & $\bf{0.18  \pm 0.02}$ & $\bf{0.37 \pm 0.03}$ & $\bf{0.57 \pm 0.02}$ & $\textcolor{blue}{0.45 \pm 0.04}$\\
\hline
Only Source NP& $0.40  \pm 0.07$ & $\textcolor{blue}{0.22   \pm 0.02}$ & $0.7   \pm 0.02$ & $0.83   \pm 0.04$ & $\bf{0.42   \pm 0.01}$\\
\hline
Pooled ST NP& $0.39  \pm 0.04$  & $\textcolor{blue}{0.22  \pm 0.01}$ & $0.72  \pm 0.04$ & $0.84  \pm 0.03$ & $\bf{0.42  \pm 0.02}$\\
\hline
TLOD& $\textcolor{blue}{0.3  \pm 0.05}$ & $0.23  \pm 0.02$ & $\textcolor{blue}{0.4  \pm 0.04}$ & $0.82  \pm 0.03$ & $0.6  \pm 0.05$ \\
\hline
Only Target NP& $\textcolor{blue}{0.29  \pm 0.05}$ & $0.26  \pm 0.02$ & $\textcolor{blue}{0.4  \pm 0.04}$ & $0.83  \pm 0.01$ & $0.6  \pm 0.05$ \\
\hline
Only Target Thr.& $0.36  \pm 0.02$ & $0.43  \pm 0.03$ & $0.6  \pm 0.09$ & $0.84  \pm 0.01$ & $0.53  \pm 0.04$ \\
\hline
Pooled ST Thr.& $0.35  \pm 0.03$ & $0.4  \pm 0.02$& $0.68  \pm 0.1$ & $0.84  \pm 0.01$& $0.53  \pm 0.04$ \\
\hline
\end{tabularx}
}
\vspace{8pt}
\caption{\small This table summarizes the Type-II errors of various approaches on different datasets for the case where $n_T=50$ and $n_S=2500$.}
\label{tab:blank_table}
\end{table}




\clearpage

\bibliographystyle{unsrt}  
\bibliography{references} 

\begin{thebibliography}{10}

\bibitem{folino2023learning}
Gianluigi Folino, Massimo Guarascio, and Francesco Chiaravalloti.
\newblock Learning ensembles of deep neural networks for extreme rainfall event
  detection.
\newblock {\em Neural Computing and Applications}, 35(14):10347--10360, 2023.

\bibitem{mazzoglio2019improving}
Paola Mazzoglio, Francesco Laio, Simone Balbo, Piero Boccardo, and Franca
  Disabato.
\newblock Improving an extreme rainfall detection system with gpm imerg data.
\newblock {\em Remote Sensing}, 11(6):677, 2019.

\bibitem{frame2022deep}
Jonathan~M Frame, Frederik Kratzert, Daniel Klotz, Martin Gauch, Guy Shalev,
  Oren Gilon, Logan~M Qualls, Hoshin~V Gupta, and Grey~S Nearing.
\newblock Deep learning rainfall--runoff predictions of extreme events.
\newblock {\em Hydrology and Earth System Sciences}, 26(13):3377--3392, 2022.

\bibitem{bourzac2014diagnosis}
Katherine Bourzac.
\newblock Diagnosis: early warning system.
\newblock {\em Nature}, 513(7517):S4--S6, 2014.

\bibitem{myszczynska2020applications}
Monika~A Myszczynska, Poojitha~N Ojamies, Alix~MB Lacoste, Daniel Neil, Amir
  Saffari, Richard Mead, Guillaume~M Hautbergue, Joanna~D Holbrook, and Laura
  Ferraiuolo.
\newblock Applications of machine learning to diagnosis and treatment of
  neurodegenerative diseases.
\newblock {\em Nature reviews neurology}, 16(8):440--456, 2020.

\bibitem{alamro2023automated}
Hayam Alamro, Wafa Mtouaa, Sumayh Aljameel, Ahmed~S Salama, Manar~Ahmed Hamza,
  and Aladdin~Yahya Othman.
\newblock Automated android malware detection using optimal ensemble learning
  approach for cybersecurity.
\newblock {\em IEEE Access}, 2023.

\bibitem{kumar2019edima}
Ayush Kumar and Teng~Joon Lim.
\newblock Edima: Early detection of iot malware network activity using machine
  learning techniques.
\newblock In {\em 2019 IEEE 5th World Forum on Internet of Things (WF-IoT)},
  pages 289--294. IEEE, 2019.

\bibitem{pan2009survey}
Sinno~Jialin Pan and Qiang Yang.
\newblock A survey on transfer learning.
\newblock {\em IEEE Transactions on knowledge and data engineering},
  22(10):1345--1359, 2009.

\bibitem{zhuang}
Fuzhen Zhuang, Zhiyuan Qi, Keyu Duan, Dongbo Xi, Yongchun Zhu, Hengshu Zhu, Hui
  Xiong, and Qing He.
\newblock A comprehensive survey on transfer learning.
\newblock {\em Proceedings of the IEEE}, 2020.

\bibitem{kalantight}
Mohammadreza~Mousavi Kalan and Samory Kpotufe.
\newblock Tight rates in supervised outlier transfer learning.
\newblock In {\em The Twelfth International Conference on Learning
  Representations}, 2024.

\bibitem{hanneke2019value}
Steve Hanneke and Samory Kpotufe.
\newblock On the value of target data in transfer learning.
\newblock {\em Advances in Neural Information Processing Systems}, 32, 2019.

\bibitem{yu2024climsim}
Sungduk Yu, Walter Hannah, Liran Peng, Jerry Lin, Mohamed~Aziz Bhouri, Ritwik
  Gupta, Bj{\"o}rn L{\"u}tjens, Justus~C Will, Gunnar Behrens, Julius Busecke,
  et~al.
\newblock Climsim: A large multi-scale dataset for hybrid physics-ml climate
  emulation.
\newblock {\em Advances in Neural Information Processing Systems}, 36, 2024.

\bibitem{nasa_power_api}
{NASA POWER}.
\newblock Nasa power: Prediction of worldwide energy resource api, 2024.

\bibitem{github_credit}
Ian Gregory.
\newblock Kaggle dataset - give me some credit, 2018.

\bibitem{uyar2010handling}
Asli Uyar, Ayse Bener, HN~Ciracy, and Mustafa Bahceci.
\newblock Handling the imbalance problem of ivf implantation prediction.
\newblock {\em IAENG International Journal of Computer Science},
  37(2):164--170, 2010.

\bibitem{abd2013review}
Shaza~M Abd~Elrahman and Ajith Abraham.
\newblock A review of class imbalance problem.
\newblock {\em Journal of Network and Innovative Computing}, 1:9--9, 2013.

\bibitem{saito2015precision}
Takaya Saito and Marc Rehmsmeier.
\newblock The precision-recall plot is more informative than the roc plot when
  evaluating binary classifiers on imbalanced datasets.
\newblock {\em PloS one}, 10(3):e0118432, 2015.

\bibitem{blitzer2007learning}
John Blitzer, Koby Crammer, Alex Kulesza, Fernando Pereira, and Jennifer
  Wortman.
\newblock Learning bounds for domain adaptation.
\newblock {\em Advances in neural information processing systems}, 20, 2007.

\bibitem{mansour2009domain}
Yishay Mansour, Mehryar Mohri, and Afshin Rostamizadeh.
\newblock Domain adaptation: Learning bounds and algorithms.
\newblock {\em arXiv preprint arXiv:0902.3430}, 2009.

\bibitem{ben2010theory}
Shai Ben-David, John Blitzer, Koby Crammer, Alex Kulesza, Fernando Pereira, and
  Jennifer~Wortman Vaughan.
\newblock A theory of learning from different domains.
\newblock {\em Machine learning}, 79(1-2):151--175, 2010.

\bibitem{ben2010impossibility}
Shai Ben-David, Tyler Lu, Teresa Luu, and D{\'a}vid P{\'a}l.
\newblock Impossibility theorems for domain adaptation.
\newblock In {\em International Conference on Artificial Intelligence and
  Statistics}, pages 129--136, 2010.

\bibitem{zhao2019learning}
Han Zhao, Remi Tachet~des Combes, Kun Zhang, and Geoffrey~J Gordon.
\newblock On learning invariant representation for domain adaptation.
\newblock {\em arXiv preprint arXiv:1901.09453}, 2019.

\bibitem{cai2021transfer}
T~Tony Cai and Hongji Wei.
\newblock Transfer learning for nonparametric classification: Minimax rate and
  adaptive classifier.
\newblock {\em The Annals of Statistics}, 2021.

\bibitem{steinwart2005classification}
Ingo Steinwart, Don Hush, and Clint Scovel.
\newblock A classification framework for anomaly detection.
\newblock {\em Journal of Machine Learning Research}, 6(2), 2005.

\bibitem{polonik1995measuring}
Wolfgang Polonik.
\newblock Measuring mass concentrations and estimating density contour
  clusters-an excess mass approach.
\newblock {\em The annals of Statistics}, pages 855--881, 1995.

\bibitem{tsybakov1997nonparametric}
Alexandre~B Tsybakov.
\newblock On nonparametric estimation of density level sets.
\newblock {\em The Annals of Statistics}, 25(3):948--969, 1997.

\bibitem{abe2006outlier}
Naoki Abe, Bianca Zadrozny, and John Langford.
\newblock Outlier detection by active learning.
\newblock In {\em Proceedings of the 12th ACM SIGKDD international conference
  on Knowledge discovery and data mining}, pages 504--509, 2006.

\bibitem{chalapathy2018anomaly}
Raghavendra Chalapathy, Aditya~Krishna Menon, and Sanjay Chawla.
\newblock Anomaly detection using one-class neural networks.
\newblock {\em arXiv preprint arXiv:1802.06360}, 2018.

\bibitem{yang2023anomaly}
Ziyi Yang, Iman Soltani, and Eric Darve.
\newblock Anomaly detection with domain adaptation.
\newblock In {\em Proceedings of the IEEE/CVF Conference on Computer Vision and
  Pattern Recognition}, pages 2958--2967, 2023.

\bibitem{andrews2016transfer}
Jerone Andrews, Thomas Tanay, Edward~J Morton, and Lewis~D Griffin.
\newblock Transfer representation-learning for anomaly detection.
\newblock {\em The Journal of Machine Learning Research}, 2016.

\bibitem{tong2018neyman}
Xin Tong, Yang Feng, and Jingyi~Jessica Li.
\newblock Neyman-pearson classification algorithms and np receiver operating
  characteristics.
\newblock {\em Science advances}, 4(2):eaao1659, 2018.

\bibitem{lehmann1986testing}
Erich~Leo Lehmann and EL~Lehmann.
\newblock {\em Testing statistical hypotheses}, volume~2.
\newblock Springer, 1986.

\bibitem{bao2020calibrated}
Han Bao, Clay Scott, and Masashi Sugiyama.
\newblock Calibrated surrogate losses for adversarially robust classification.
\newblock In {\em Conference on Learning Theory}, pages 408--451. PMLR, 2020.

\bibitem{bartlett2002rademacher}
Peter~L Bartlett and Shahar Mendelson.
\newblock Rademacher and gaussian complexities: Risk bounds and structural
  results.
\newblock {\em Journal of Machine Learning Research}, 3(Nov):463--482, 2002.

\bibitem{golowich2018size}
Noah Golowich, Alexander Rakhlin, and Ohad Shamir.
\newblock Size-independent sample complexity of neural networks.
\newblock In {\em Conference On Learning Theory}, pages 297--299. PMLR, 2018.

\bibitem{saidi2015assessment}
Helmi Saidi, Marzia Ciampittiello, Claudia Dresti, and Giorgio Ghiglieri.
\newblock Assessment of trends in extreme precipitation events: a case study in
  piedmont (north-west italy).
\newblock {\em Water Resources Management}, 29:63--80, 2015.

\bibitem{schar2016percentile}
Christoph Sch{\"a}r, Nikolina Ban, Erich~M Fischer, Jan Rajczak, J{\"u}rg
  Schmidli, Christoph Frei, Filippo Giorgi, Thomas~R Karl, Elizabeth~J Kendon,
  Albert MG~Klein Tank, et~al.
\newblock Percentile indices for assessing changes in heavy precipitation
  events.
\newblock {\em Climatic Change}, 137:201--216, 2016.

\bibitem{kim2023exposure}
Jungho Kim, Jeremy Porter, and Edward~J Kearns.
\newblock Exposure of the us population to extreme precipitation risk has
  increased due to climate change.
\newblock {\em Scientific reports}, 13(1):21782, 2023.

\bibitem{koltchinskii2011oracle}
Vladimir Koltchinskii.
\newblock Oracle inequalities in empirical risk minimization and sparse
  recovery problems, volume 2033 of lecture notes in mathematics, 2011.

\bibitem{shalev2014understanding}
Shai Shalev-Shwartz and Shai Ben-David.
\newblock {\em Understanding machine learning: From theory to algorithms}.
\newblock Cambridge university press, 2014.

\bibitem{rigollet2011neyman}
Philippe Rigollet and Xin Tong.
\newblock Neyman-pearson classification, convexity and stochastic constraints.
\newblock {\em Journal of machine learning research}, 2011.

\end{thebibliography}

\appendix

\section{Proof of Proposition 1}

By [\cite{koltchinskii2011oracle}, Theorem 2.3.], we can get $R_n(\varphi \circ \mathcal{H})\leq 2L R_n(\mathcal{H})$. Then, applying McDiarmid's inequality [See \cite{shalev2014understanding}, Chapter 26], we obtain 
\begin{align*}
\underset{h\in \mathcal{H}}{\sup}\ \bigg|\mathbb{E}_{\mu}[\varphi(h(X))]-\frac{1}{n}\underset{X_i\sim \mu}{\Sigma}\varphi(h(X_i))\bigg|&\leq 2R_n(\varphi \circ \mathcal{H})+ C\sqrt{\frac{2\log(2/\delta)}{n}}\\
&\leq \frac{4LB_{\mathcal{H}}}{\sqrt{n}}+C\sqrt{\frac{2\log(2/\delta)}{n}}
\end{align*}
with probability at least $1-\delta$. Furthermore, since $R_n(\mathcal{H})=R_n(\mathcal{H}^{-})$, where $\mathcal{H}^{-}=\{-h:h\in \mathcal{H}\}$, the same bound applies to the expression $$\underset{h\in \mathcal{H}}{\sup}\ \bigg|\mathbb{E}_{\mu}[\varphi(-h(X))]-\frac{1}{n}\underset{X_i\sim \mu}{\Sigma}\varphi(-h(X_i))\bigg|,$$
which concludes Proposition 1.
\begin{flushright}{$\square$}
\end{flushright}

\section{Proof of Theorem 1}
Consider the event where Proposition 1 holds for the distributions $\mu_0,\mu_{1,S}$, and $\mu_{1,T}$, which occurs with probability at least $1-3\delta$. We then divide the proof into three parts: 1)$R_{\varphi,\mu_0}(\hat{h})\leq \alpha+\epsilon_0$, 2) $\mathcal{E}_{1,T}(\hat{h})\leq \frac{4\tilde{C}}{\sqrt{n_T}}$, 3) $\mathcal{E}_{1,T}(\hat{h})\leq c_{\rho(r)}\cdot( \frac{\tilde{C}}{\sqrt{n_S}})^{1/\rho(r)}+4\cdot\Delta$

\textbf{Part 1)} $R_{\mu_0}(\hat{h})\leq R_{\varphi,\mu_0}(\hat{h})\leq \alpha+\epsilon_0$. The first inequality holds because $\varphi$ is non-decreasing and $\varphi(0)=1$. Moreover, due to Proposition 1 for the distribution $\mu_0$ and the constraint on $\hat{R}_{\varphi,\mu_0}(\hat{h})$ in (8), we obtain 
$$R_{\varphi,\mu_0}(\hat{h})\leq \hat{R}_{\varphi,\mu_0}(\hat{h})+\epsilon_0/2\leq \alpha+\epsilon_0.$$

\textbf{Part 2)} $\mathcal{E}_{1,T}(\hat{h})\leq \frac{4\tilde{C}}{\sqrt{n_T}}$. Note that since $R_{\varphi,\mu_0}(h^*_{T,\alpha})\leq \alpha$, Proposition 1 gives us $\hat{R}_{\varphi,\mu_0}(h^*_{T,\alpha})\leq \alpha +\epsilon_0/2$. Therefore, $h^*_{T,\alpha}$ belongs to the constraint set in the optimization problem (7), which implies that $\hat{R}_{\varphi,\mu_{1,T}}(\hat{h}_{T,\alpha+\epsilon_0/2})\leq\hat{R}_{\varphi,\mu_{1,T}}(h^*_{T,\alpha})$. Then, we can get

\begin{align*}
R_{\varphi,\mu_{1,T}}(\hat{h})-R_{\varphi,\mu_{1,T}}(h^*_{T,\alpha})&\leq \hat{R}_{\varphi,\mu_{1,T}}(\hat{h})-\hat{R}_{\varphi,\mu_{1,T}}(h^*_{T,\alpha})+\frac{\tilde{C}}{\sqrt{n_T}}\\
&\leq \hat{R}_{\varphi,\mu_{1,T}}(\hat{h}_{T,\alpha+\epsilon_0/2})-\hat{R}_{\varphi,\mu_{1,T}}(h^*_{T,\alpha})+\frac{4\tilde{C}}{\sqrt{n_T}}\leq \frac{4\tilde{C}}{\sqrt{n_T}}
\end{align*}

where the first inequality follows from Proposition 1, and the second uses the constraint on $\hat{R}_{\varphi,\mu_{1,T}}(\hat{h})$ in (8).

\textbf{Part 3)} $\mathcal{E}_{1,T}(\hat{h})\leq c_{\rho(r)}\cdot( \frac{\tilde{C}}{\sqrt{n_S}})^{1/\rho(r)}+4\cdot\Delta$. First, we define $\tilde{h}_{T,\alpha+\epsilon_0/2}$ as follows:

\begin{align*}
\tilde{h}_{T,\alpha+\epsilon_0/2}=&\argmin_{h\in \mathcal{H}} \ R_{\varphi,\mu_{1,T}}(h)\nonumber\\
    & \text{s.t.} \ \hat{R}_{\varphi,\mu_0}(h)\leq \alpha+\epsilon_0/2.
\end{align*}
Note that we have 
\begin{align}\label{3_inequalities}
R_{\varphi,\mu_{1,T}}(h^*_{T,\alpha+\epsilon_0})\leq R_{\varphi,\mu_{1,T}}(\tilde{h}_{T,\alpha+\epsilon_0/2})\leq R_{\varphi,\mu_{1,T}}(\hat{h}_{T,\alpha+\epsilon_0/2}).
\end{align}
We then divide the proof into two cases:

\textbf{Part 3, Case I:} If $R_{\varphi,\mu_{1,T}}(h^*_{S,\alpha})-R_{\varphi,\mu_{1,T}}(\tilde{h}_{T,\alpha+\epsilon_0/2})>\frac{\tilde{C}}{\sqrt{n_T}}$. Then, \eqref{3_inequalities} implies that $\Delta>\frac{\tilde{C}}{\sqrt{n_T}}$, where $\Delta$ is defined as $\Delta=R_{\varphi,\mu_{1,T}}(h^*_{S,\alpha})-R_{\varphi,\mu_{1,T}}(h^*_{T,\alpha+\epsilon_0})$ in Theorem 1. Therefore, the inequality \linebreak$\mathcal{E}_{1,T}(\hat{h})\leq c_{\rho(r)}\cdot( \frac{\tilde{C}}{\sqrt{n_S}})^{1/\rho(r)}+4\cdot\Delta$ becomes trivial due to part 2.

\textbf{Part 3, Case II:} If $R_{\varphi,\mu_{1,T}}(h^*_{S,\alpha})-R_{\varphi,\mu_{1,T}}(\tilde{h}_{T,\alpha+\epsilon_0/2})\leq\frac{\tilde{C}}{\sqrt{n_T}}$. We first claim that $h^*_{S,\alpha}$ belongs to the constraint set in (8). To show that, we have 
$$R_{\varphi,\mu_{1,T}}(h^*_{S,\alpha})\leq R_{\varphi,\mu_{1,T}}(\tilde{h}_{T,\alpha+\epsilon_0/2})+\frac{\tilde{C}}{\sqrt{n_T}}\leq R_{\varphi,\mu_{1,T}}(\hat{h}_{T,\alpha+\epsilon_0/2})+\frac{\tilde{C}}{\sqrt{n_T}}\leq \hat{R}_{\varphi,\mu_{1,T}}(\hat{h}_{T,\alpha+\epsilon_0/2})+\frac{3\tilde{C}}{2\sqrt{n_T}}.$$

Furthermore, we can obtain 
$$\hat{R}_{\varphi,\mu_{1,T}}(h^*_{S,\alpha})\leq R_{\varphi,\mu_{1,T}}(h^*_{S,\alpha})+\frac{\tilde{C}}{2\sqrt{n_T}}\leq \hat{R}_{\varphi,\mu_{1,T}}(\hat{h}_{T,\alpha+\epsilon_0/2})+\frac{2\tilde{C}}{\sqrt{n_T}},$$
which implies that $h^*_{S,\alpha}$ belongs to the constraint set in (8). Hence, we get
$$R_{\varphi,\mu_{1,S}}(\hat{h})-R_{\varphi,\mu_{1,S}}(h^*_{S,\alpha})\leq \hat{R}_{\varphi,\mu_{1,S}}(\hat{h})-\hat{R}_{\varphi,\mu_{1,S}}(h^*_{S,\alpha})+\frac{\tilde{C}}{\sqrt{n_S}}\leq \frac{\tilde{C}}{\sqrt{n_S}}.$$

Then, since $R_{\varphi,\mu_0}(\hat{h})\leq \alpha+\epsilon_0\leq \alpha+r$, by Definition 5 we obtain
$$R_{\varphi,\mu_{1,T}}(\hat{h})-R_{\varphi,\mu_{1,T}}(h^*_{S,\alpha})\leq c_{\rho(r)}\cdot( \frac{\tilde{C}}{\sqrt{n_S}})^{1/\rho(r)}.$$
Therefore,
\begin{align*}
    R_{\varphi,\mu_{1,T}}(\hat{h})-R_{\varphi,\mu_{1,T}}(h^*_{T,\alpha})&\leq R_{\varphi,\mu_{1,T}}(\hat{h})-R_{\varphi,\mu_{1,T}}(\tilde{h}_{T,\alpha+\epsilon_0/2})\\
    &=R_{\varphi,\mu_{1,T}}(\hat{h})-R_{\varphi,\mu_{1,T}}(h^*_{S,\alpha})+R_{\varphi,\mu_{1,T}}(h^*_{S,\alpha})-R_{\varphi,\mu_{1,T}}(\tilde{h}_{T,\alpha+\epsilon_0/2})\\
    &\leq c_{\rho(r)}\cdot( \frac{\tilde{C}}{\sqrt{n_S}})^{1/\rho(r)}+R_{\varphi,\mu_{1,T}}(h^*_{S,\alpha})-R_{\varphi,\mu_{1,T}}(h^*_{T,\alpha+\epsilon_0})\\
    &\leq  c_{\rho(r)}\cdot( \frac{\tilde{C}}{\sqrt{n_S}})^{1/\rho(r)}+4\cdot \Delta.
\end{align*}
\begin{flushright}{$\square$}
\end{flushright}
\section{Proof of Theorem 2}
We use some ideas from the proof of Proposition 4.1 in \cite{rigollet2011neyman}. First, we show that $\gamma(\alpha):=\inf_{h_{\theta}\in \mathcal{H}_{\alpha}(\mu_0)} R_{\varphi,\mu_{1,T}}(h_{\theta})$ is a non-increasing convex function on $[0,1]$, where $\mathcal{H}_{\alpha}(\mu_0)=\{h_{\theta}\in \mathcal{H}: R_{\varphi,\mu_0}(h_{\theta})\leq \alpha\}$. The non-increasing property is straightforward to verify.

Next, we take $\alpha_1,\alpha_2 \in [0,1]$ and aim to show that for any $\theta\in (0,1)$, the following inequality holds: \begin{align}\label{gamma_inequality}
\gamma(\bar{\alpha})\leq \theta\gamma(\alpha_1)+(1-\theta)\gamma(\alpha_2).
\end{align}
where $\bar{\alpha}=\theta\alpha_1+(1-\theta)\alpha_2$. Let $\epsilon>0$ be an arbitrary small number. Then, there exist $h_1\in \mathcal{H}_{\alpha_1}(\mu_0)$ and $h_2\in \mathcal{H}_{\alpha_2}(\mu_0)$ such that $R_{\varphi,\mu_{1,T}}(h_1)\leq \gamma(\alpha_1)+\epsilon$ and $R_{\varphi,\mu_{1,T}}(h_2)\leq \gamma(\alpha_2)+\epsilon$. Consider the convex combination $h_3=\theta\cdot h_1+(1-\theta)\cdot h_2$, which by assumption belongs to $\mathcal{H}$ . By the convexity of $\varphi$ we have $$R_{\varphi,\mu_0}(h_3)\leq \theta R_{\varphi,\mu_0}(h_1)+(1-\theta)R_{\varphi,\mu_0}(h_2)\leq \bar{\alpha},$$ This implies that $h_3\in \mathcal{H}_{\bar{\alpha}}(\mu_0)$. Therefore, $$\gamma(\bar{\alpha})\leq R_{\varphi,\mu_{1,T}}(h_3)\leq \theta R_{\varphi,\mu_{1,T}}(h_1)+(1-\theta)R_{\varphi,\mu_{1,T}}(h_2)\leq \theta \gamma(\alpha_1)+(1-\theta)\gamma(\alpha_2)+\epsilon.$$
Since $\epsilon>0$ is arbitrary, we conclude that the inequality \eqref{gamma_inequality} holds, which implies that


$$\gamma(\alpha)-\gamma(\alpha+\epsilon_0)\leq \epsilon_0 \frac{\gamma(\alpha/2)-\gamma(\alpha)}{\alpha/2}\leq \frac{C\epsilon_0}{\alpha/2}.$$ 

Then, we can bound the following term:
\begin{align*}
    R_{\varphi,\mu_{1,T}}(h^*_{T,\alpha})-R_{\varphi,\mu_{1,T}}(h^*_{T,\alpha+\epsilon_0})=\gamma(\alpha)-\gamma(\alpha+\epsilon_0)\leq \frac{2C\tilde{C}}{\alpha\sqrt{n_0}}.
\end{align*}
\begin{flushright}{$\square$}
\end{flushright}

\section{More Details on the Climate Dataset \citep{yu2024climsim}}\label{appendix_D}

In Section \ref{sec:climate_data_climsim}, we used different location clusters as pairs of source and target. The original dataset \citep{yu2024climsim} includes various locations specified by longitude and latitude, as shown in Figure \ref{fig:locations}. Since each location does not have sufficient data for creating training and test samples, we group neighboring locations to form clusters, as illustrated in Figure \ref{fig:second_figure}. These clustered locations are then used as source and target pairs.

\begin{figure}[h]
    \centering
    \begin{subfigure}[b]{0.47\textwidth} 
        \centering
        \includegraphics[width=\textwidth]{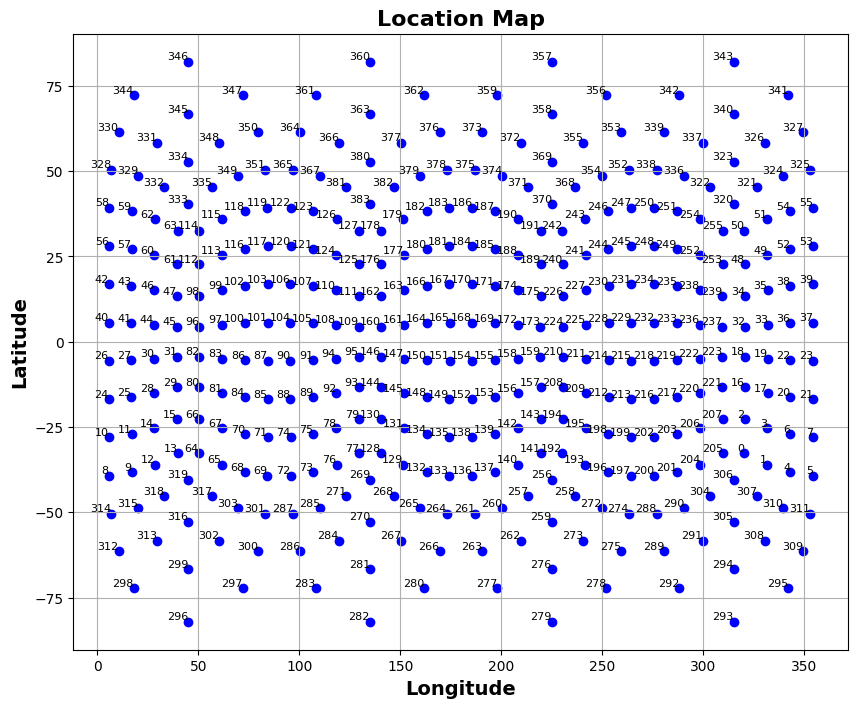} 
        \caption{Various locations where climate data has been recorded.}
        \label{fig:locations}
    \end{subfigure}
    \hfill 
    \begin{subfigure}[b]{0.5\textwidth} 
        \centering
        \includegraphics[width=\textwidth]{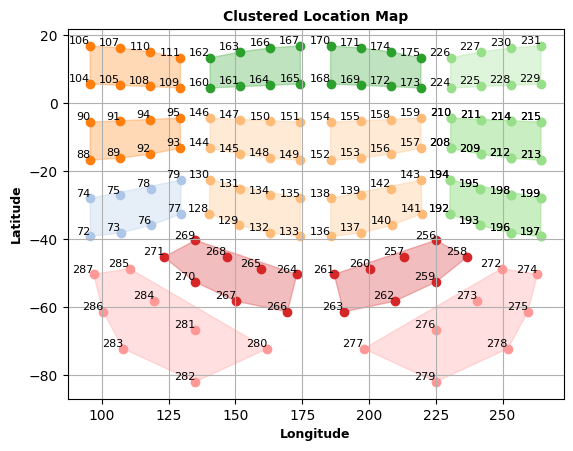} 
        \caption{Clustered locations by grouping neighboring ones.}
        \label{fig:second_figure}
    \end{subfigure}
    \caption{}
    \label{fig:clusters}
\end{figure}

\section{Alternative Approach to Filter $\hat{\mathcal{H}}$ in Step 2 of TLNP (Section \ref{sec:transfer_learning_algorithm})}\label{variance_method_appendix}
In Section \ref{sec:transfer_learning_algorithm}, in Step 2, we use a universal constant $c=0.5$ in the inequality \eqref{step_2_thresholding} to filter the functions in $\hat{\mathcal{H}}$. As the constant serves primarily to upper-bound the variance of errors for a given dataset, we propose an alternative approach here by estimating the variance as follows.

First, we divide the target abnormal data into 70\% for training and 30\% for evaluation. Let $n_T$ represent the number of
data points in the training set. Step 1 is the same as the procedure described in Section \ref{sec:transfer_learning_algorithm}. 

In  Step 2, we first repeat Step 1 using the $30\%$ of the target abnormal training data set aside for evaluation, along with all data from the normal class, i.e., $\mu_0$, and without using any source data, i.e., $\lambda_S = 0$. This process yields a function $\hat{h}_T \in \mathcal{H}$. Inspired by the constraint in the optimization procedure \eqref{algorithm_main}, we filter the functions in $\hat{\mathcal{H}}$, obtained in the first step using $n_T$ target abnormal training data, by comparing their performance with that of $\hat{h}_T$ as follows. First, we calculate the output of $\text{sign}(\hat{h}_T)$ on the $n_T$ target abnormal training data and compute the variance of the resulting $\pm 1$ outputs, denoted as VAR. Let $\hat{R}_{\mu_{1,T}}$ represent the target $0$-$1$ loss (Type-II error) computed with respect to the $n_T$ target abnormal training data. We then define $\hat{\mathcal{H}}_T$ as the set of functions $h \in \hat{\mathcal{H}}$ that satisfy the following inequality:
\begin{align*}
\hat{R}_{\mu_{1,T}}(\text{sign}(h))\leq \hat{R}_{\mu_{1,T}}(\text{sign}(\hat{h}_T))+\sqrt{\frac{\text{VAR}}{n_T}}
\end{align*}

Step 3 remains the same as the one described in Section \ref{sec:transfer_learning_algorithm}.

In Figures \ref{variance_method_38,37} and \ref{fig:credit_variance}, we demonstrate the results obtained using this approach, referred to as the TLNP variance method, and compare it with the procedure described in Section \ref{sec:transfer_learning_algorithm}. The results indicate that both methods yield nearly the same performance. In Figure \ref{variance_method_38,37}, TLNP and TLNP varaince method are identical and overlap completely. In Figure \ref{fig:credit_variance}, when $n_T$ is sufficiently large, TLNP slightly outperforms TLNP variance method; however, both methods outperform other approaches.

\begin{figure}[h]
    \centering
    \begin{tikzpicture}
        \begin{axis}[
            title={Heavy Rain  (Climsim), Target and Source Clusters: 38 and 37,39, $n_T=50$ },xlabel={Number of Abnormal Source Samples},
            ylabel={Test Target Type-II Error},
            legend pos=north west, ymin=0.1, ymax=0.5,
            grid=major,label style={font=\small},
            width=.7\columnwidth, height=4.5cm,
            legend style={at={(0.5,-.45)}, anchor=north, legend columns=1, font=\footnotesize},
            ]
            
\addplot[
    mark=o,
    thick,
    color=orange
] coordinates {
    (100, 0.231)
    (500, 0.310)
    (1000, 0.187)
    (2000, 0.185)
    (2500, 0.21827)
};
\addlegendentry{Only Source Neyman-Pearson}

\addplot[
    mark=o,
    thick,
    color=brown
] coordinates {
    (100, 0.22662)
    (500, 0.2629)
    (1000, 0.1876)
    (2000, 0.17864)
    (2500, 0.2188)
};
\addlegendentry{Pooled Source and Target Neyman-Pearson}

\addplot[
    mark=o,
    thick,
    color=red
] coordinates {
    (100, 0.4069)
    (500, 0.386)
    (1000, 0.4155)
    (2000, 0.400)
    (2500, 0.400)
};
\addlegendentry{Pooled Source and Target Thresholding Traditional Classification}

\addplot[
    mark=o,
    ultra thick,
    color=black
] coordinates {
     (100, 0.19559)
    (500, 0.1909)
    (1000, 0.1787)
    (2000, 0.1781)
    (2500, 0.177)
};
\addlegendentry{TLNP}

\addplot[
    mark=o,
    ultra thick,
    opacity=0.6,
    color=magenta!70,
] coordinates {
     (100, 0.19559)
    (500, 0.1909)
    (1000, 0.1787)
    (2000, 0.1781)
    (2500, 0.177)
};
\addlegendentry{TLNP Variance Method}

\addplot[
    mark=o,
    thick,
    opacity=0.7,
    color=green!50!black,
] coordinates {
    (100, 0.2439)
    (500, 0.265356)
    (1000, 0.204)
    (2000, 0.193)
    (2500, 0.2304)
};
\addlegendentry{Transfer Learning Outlier Detection}

\addplot[
    mark=o,
    thick,
    dashed,
    color=blue
] coordinates {
    (100, 0.2638)
    (500, 0.2638)
    (1000, 0.2638)
    (2000, 0.2638)
    (2500, 0.2638)
};
\addlegendentry{Only Target Neyman-Pearson}
\addplot[
    mark=star,
    thick,
    dashed,
    color=purple!70!black
] coordinates {
    (100, 0.427)
    (500, 0.427)
    (1000, 0.427)
    (2000, 0.427)
    (2500, 0.427)
};
\addlegendentry{Only Target Thresholding Traditional Classification}

        \end{axis}
    \end{tikzpicture}
    \caption{\small The performance of the TLNP variance method, along with other approaches described in Section \ref{sec:transfer_learning_algorithm}. The threshold for Type-I error is set at $\alpha = 0.05$, and the experimental settings, including the number of samples, are identical to those in Figure \ref{fig:two_plots_with_shared_legend_38,37,39}.
.
}
    \label{variance_method_38,37} 
\end{figure}
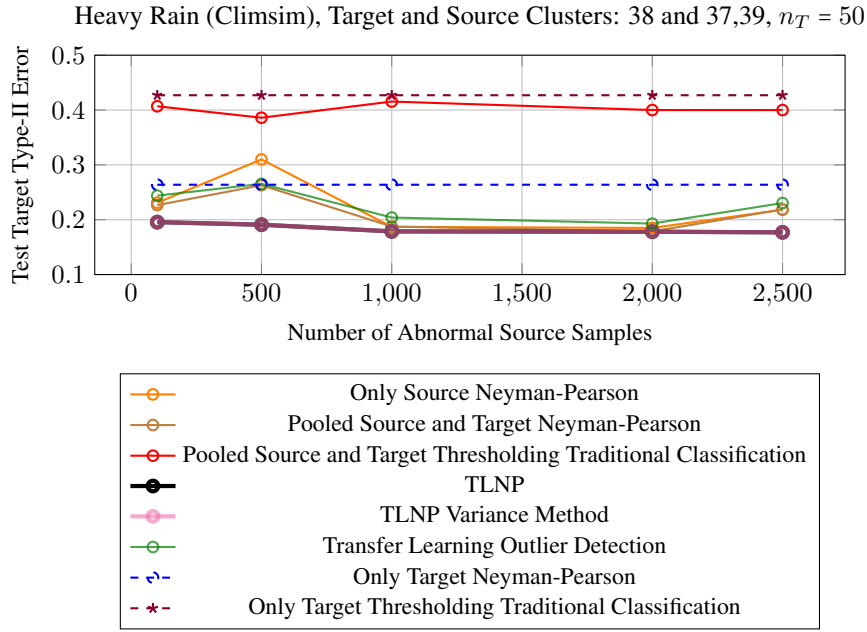

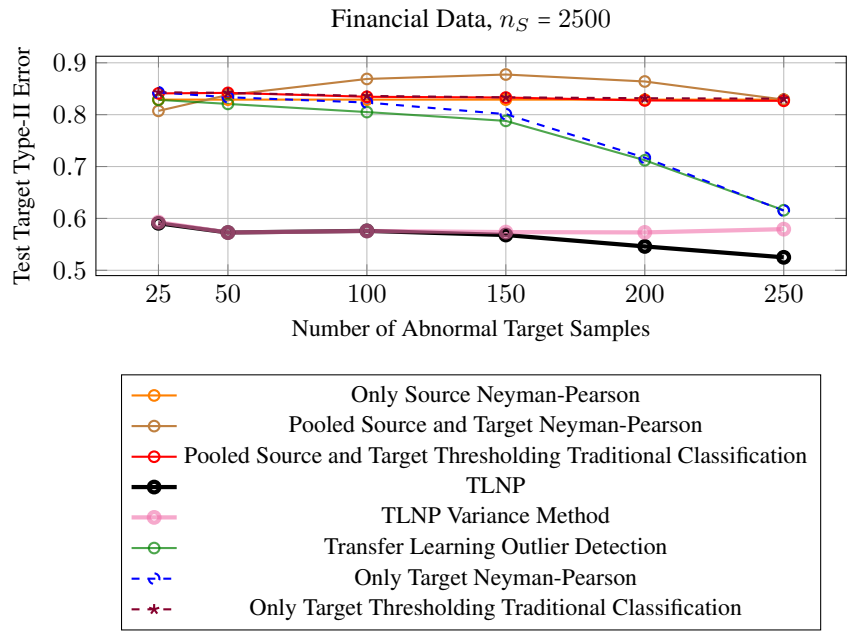
\begin{figure}[h]
    \centering
    \begin{tikzpicture}
        \begin{axis}[
            title={Financial Data, $n_S=2500$},xlabel={Number of Abnormal Target Samples},
            ylabel={Test Target Type-II Error},
            legend pos=north west,
            grid=major,
            xtick={25, 50, 100, 150, 200,250},
            label style={font=\small},
            width=.7\columnwidth, height=4.5cm,
            legend style={at={(0.5,-.45)}, anchor=north, legend columns=1, font=\footnotesize},
            ]

\addplot[
    mark=o,
    thick,
    color=orange
] coordinates {
    (25, 0.829)
    (50, 0.829)
    (100, 0.829)
    (150, 0.829)
    (200, 0.829)
    (250, 0.829)
};
\addlegendentry{Only Source Neyman-Pearson}

\addplot[
    mark=o,
    thick,
    color=brown
] coordinates {
    (25, 0.8073)
    (50, 0.838)
    (100, 0.86884)
    (150, 0.8774)
    (200, 0.864)
    (250, 0.829)
};
\addlegendentry{Pooled Source and Target Neyman-Pearson}

\addplot[
    mark=o,
    thick,
    color=red
] coordinates {
    (25, 0.841)
    (50, 0.842)
    (100, 0.8345)
    (150, 0.8330)
    (200, 0.8274)
    (250, 0.8267)
};
\addlegendentry{Pooled Source and Target Thresholding Traditional Classification}

\addplot[
    mark=o,
    ultra thick,
    color=black
] coordinates {
    (25, 0.5906)
    (50, 0.5728)
    (100, 0.576)
    (150, 0.568)
    (200, 0.546)
    (250, 0.525)
};
\addlegendentry{TLNP}

\addplot[
    mark=o,
    ultra thick,
    opacity=0.6,
    color=magenta!70,
] coordinates {
    (25, 0.5928)
    (50, 0.5728)
    (100, 0.5758)
    (150, 0.57363)
    (200, 0.573)
    (250, 0.5793)
};
\addlegendentry{TLNP Variance Method}

\addplot[
    mark=o,
    thick,
    opacity=0.7,
    color=green!50!black
] coordinates {
    (25, 0.8286)
    (50, 0.8207)
    (100, 0.805)
    (150, 0.788)
    (200, 0.7121)
    (250, 0.6158)
};
\addlegendentry{Transfer Learning Outlier Detection}

\addplot[
    mark=o,
    thick,
    dashed,
    color=blue
] coordinates {
    (25, 0.843)
    (50, 0.8336)
    (100, 0.8234)
    (150, 0.801)
    (200, 0.7170)
    (250, 0.615)
};
\addlegendentry{Only Target Neyman-Pearson}

\addplot[
    mark=star,
    thick,
    dashed,
    color=purple!70!black
] coordinates {
    (25, 0.8432)
    (50, 0.842)
    (100, 0.8361)
    (150, 0.83342)
    (200, 0.8317)
    (250, 0.8308)
};
\addlegendentry{Only Target Thresholding Traditional Classification}

        \end{axis}
    \end{tikzpicture}
    \caption{\small The performance of the TLNP variance method, along with other approaches described in Section \ref{sec:transfer_learning_algorithm}. The threshold for Type-I error is set at $\alpha = 0.05$, and the experimental settings, including the number of samples, are identical to those in Figure \ref{fig:credit}}
    \label{fig:credit_variance} 
\end{figure}

\end{document}